\documentclass[11pt]{article}

\usepackage[final]{acl}

\usepackage{times}
\usepackage{latexsym}
\usepackage{booktabs}
\usepackage{subcaption}
\usepackage{microtype}
\usepackage{multirow}
\usepackage{nert}
\usepackage{tipa}
\usepackage{tikz-dependency}
\usepackage{placeins}

\depstyle{label style}{draw, fill=white, rounded corners=2pt, inner sep=1.5pt, font=\scriptsize\sffamily}
\usepackage[T1]{fontenc}

\usepackage[utf8]{inputenc}

\usepackage{microtype}

\usepackage{inconsolata}

\usepackage{graphicx}

\usepackage{graphicx}
\usepackage{titling}

\newcommand{\kite}{\raisebox{-0.2em}{\includegraphics[height=1em]{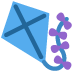}}}

\newcommand{\cait}{CAIT\ \kite}

\title{CAIT: A Syntactic Parsing Toolkit for Child--Adult InTeractions }

\author{
  \textbf{Francesca Padovani\textsuperscript{1}},  \textbf{Xiulin Yang\textsuperscript{2}},
 \textbf{Bastian Bunzeck\textsuperscript{3}},
  \textbf{Jaap Jumelet\textsuperscript{1}},
\\
  \textbf{Yevgen Matusevych\textsuperscript{1}},
  \textbf{Nathan Schneider\textsuperscript{2}},
  \textbf{Arianna Bisazza\textsuperscript{1}}
\\
\\
  \textsuperscript{1}Center for Language and Cognition (CLCG), University of Groningen, \\
  \textsuperscript{2}Georgetown University,\\
  \textsuperscript{3}Computational Linguistics, Department of Linguistics, Bielefeld University
\\
  \small{
    \textbf{Correspondence:} \href{mailto:}{f.padovani@rug.nl}
  }
}

\begin{document}
\maketitle
\begin{abstract}
CHILDES is a paramount resource for language acquisition studies---yet computational tools for analyzing its syntactic structure remain limited. Leveraging the recent release of the UD-English-CHILDES treebank with gold-standard Universal Dependencies (UD) annotations, we train a state-of-the-art dependency parser specifically tailored to CHILDES. The parser more accurately captures syntactic patterns in child--adult interactions, outperforming widely used off-the-shelf English parsers, including SpaCy and Stanza.
Alongside the parser, we also release a Part-of-Speech tagger and an utterance-level construction tagger, which together form the open-source Syntactic Parsing Toolkit for Child--Adult InTeractions (\cait). Through a detailed error analysis and a case study tracking the distribution of syntactic constructions across developmental time in CHILDES, we demonstrate the practical utility of the toolkit for large-scale, reproducible research on language acquisition\footnote{Models and data can be found at \url{https://github.com/fpadovani/CAIT-Toolkit/}}.

\end{abstract}

\section{Introduction}

The kind of language that children produce and hear in their environment differs from regular everyday language \citep[e.g.,][]{saxton2009inevitability}. Child Speech (CS) starts out with semantically motivated isolated words (e.g., \textit{there}), holistic phrases (e.g., \textit{all-gone}), and multiword utterances revolving around `pivot words' \cite{braine1976children} with an open slot (e.g., \textit{``More \_''}) (cf. \citealp{lieven1997lexicallybased,lieven2003early,tomasello2000itembased,tomasello2003constructing}). Child-Directed Speech (CDS) is similarly characterized by short, syntactically simple utterances, high lexical and structural repetition, and cross-turn redundancy \cite{LESTER2022104986, cameron-faulkner2003construction}. Both registers challenge standard linguistic analysis through non-canonical constructions, disfluencies, self-repair, and fragmentary utterances \cite{liu-prudhommeaux-2023-data}. As usage-based theories propose that linguistic competence emerges from repeated exposure to structured input patterns \cite{cameron-faulkner2003construction, lubetich2014data, you2021child}, CS and CDS are central empirical domains: the former reflects the developing linguistic system, the latter reveals the distributional patterns from which it emerges.

Many language acquisition studies rely on transcribed corpora such as CHILDES \footnote{Throughout the paper, we use the term CHILDES as a shorthand for the English child--adult interaction domain represented in the CHILDES repository and used in our experiments. When relevant, we differentiate between Child Speech (CS) and Child-Directed Speech (CDS).} \citep[e.g.,][]{macwhinney2000childes,legate2002empirical,lidz2003infants}. More recently, advances in neural language models have renewed interest in using computational methods to test acquisition hypotheses under controlled, naturalistic input conditions \citep[e.g.,][]{warstadt2022artificial,portelance2024roles, bunzeck-etal-2025-construction}. Despite its empirical and theoretical importance, CHILDES is difficult to analyze at scale. Existing computational tools struggle with its domain-specific idiosyncrasies, forcing researchers either to manually inspect data or rely on general-purpose parsers trained on written, adult-directed text---tools not designed for spoken, developmental language whose errors require costly post-hoc correction.



Most existing dependency parsers are trained on adult-directed written text (e.g., news, books, encyclopedic articles) and consequently perform poorly when applied to spoken, interactional, and developmental input such as CHILDES. While a small number of studies have explored dependency parsing for adult conversational speech  \citep[e.g.,][]{bechet2014adapting, davidson-etal-2019-dependency}, no clear consensus exists 
on a parser that can annotate CHILDES data reliably.

The recent publication of UD-English-CHILDES \cite{yang-etal-2025-ud} provides the first officially validated Universal Dependencies (UD) v2 treebank \cite{zeman-2021-universal} for English child--adult interactions. It consolidates several earlier annotation efforts \citep{liu2021dependency, liu-prudhommeaux-2023-data, szubert2025cross} into a unified resource that conforms to current UD guidelines. Building on this resource, our work presents four main contributions:
\begin{itemize}
    \item We \textbf{train and evaluate several dependency parsers} on UD-English-CHILDES \cite{yang-etal-2025-ud} under different training setups and assess their performance on a CHILDES test set. 
    \item We release \textbf{\cait} (pronounced \textipa{[kaIt]}), an open-source syntactic parsing toolkit for Child--Adult InTeractions, whose central component is our best performing \textbf{dependency parser} based on \texttt{SuPar} \cite{zhang-etal-2020-efficient}. The suite also includes a \textbf{POS tagger} trained using \texttt{Stanza} \cite{qi-etal-2020-stanza}, and an utterance-level \textbf{construction tagger} leveraging the parser annotations.

    \item We perform an \textbf{in-depth error analysis}, comparing CAIT annotations with those of the off-the-shelf English \texttt{Stanza} parser, highlighting systematic differences in dependency predictions and the structural biases introduced by general-purpose parsers.

    \item We demonstrate the practical utility of CAIT through a \textbf{case study}, in which the construction tagger is used to track the distribution of syntactic structures across development.
\end{itemize}

\noindent
Overall, our results show that an in-domain CHILDES model can enhance parsing accuracy on CS and CDS, offering a valuable tool for both computational modeling that requires strict syntactic manipulation of CHILDES and empirical studies of language development that rely on large-scale syntactic analysis.

\section{Related Work}
\label{sec:related-work}

\paragraph{Parsing Spoken Language}
Over the past decades, several studies have focused on developing treebanks for dialogue and conversational data \cite{caines-etal-2017-parsing,dobrovoljc-martinc-2018-er,braggaar-van-der-goot-2021-challenges}. A classic example is the pivotal work on the Switchboard \cite{godfrey1992switchboard} section of the Penn Treebank \cite{marcus-etal-1993-building}, which consists of transcribed adult--adult telephone conversations. More recent efforts have moved towards UD-style treebanks 
\cite{kahane-etal-2021-annotation,dobrovoljc-2022-spoken}, which have become the dominant syntactic
annotation framework in NLP. For instance, \citet{dobrovoljc-nivre-2016-universal}, \citet{ovrelid-etal-2018-lia} and \citet{kahane-etal-2021-annotation} introduced UD dependency annotations for adult conversational speech in Slovenian, Norwegian, and French, respectively.
On the parsing side, models trained or fine-tuned on domain-specific spoken data have been shown to outperform those trained on written text alone, e.g., \citet{bechet2014adapting} on French and \citet{davidson-etal-2019-dependency} on English human–machine dialogues, highlighting the importance of in-domain data for parsing spoken language.


\paragraph{Parsing CHILDES}
The CHILDES database \cite{macwhinney2000childes} provides dependency annotations originally developed by \citet{sagae-etal-2004-adding, sagae-etal-2005-automatic, sagae-etal-2007-high}, which, however, do not conform to the UD framework.

With UD-style annotation gaining importance in Natural Language Processing (NLP) and language acquisition research, NLP toolkits such as \texttt{Stanza} \citep{qi-etal-2020-stanza} have been used to parse CHILDES data \cite{liu2024morphosyntactic}. However, these automatic annotations remain inconsistent and unreliable.
\citet{liu-prudhommeaux-2023-data} present the first extensive UD-style dependency annotations for child--adult interactions. Their work extends earlier, smaller-scale studies that applied UD standards to this data \cite{liu2021dependency}. In their treebank, \citet{liu-prudhommeaux-2023-data} address all critical aspects of spoken dialogue, such as speech repairs, restarts, and disfluent fragments, ensuring that the resource reflects the full range of phenomena characteristic of CHILDES. The study also presents the most recent and extensive evaluation of parsers on CHILDES, testing multiple out-of-domain models on child--adult interactions. These experiments reveal that general-purpose parsers struggle with constructions typical of this domain. Training on a small set of nine annotated child--adult interactions improves parsing accuracy, particularly for younger children, though the gains over out-of-domain models remain modest. Despite being a useful resource, their dataset is not fully consistent with UD v2, lacks UPOS tags, and has not undergone independent validation. 

To address these limitations, \citet{yang-etal-2025-ud} have recently undertaken a major harmonization effort, integrating and expanding previous datasets to produce a unified UD v2–compliant resource for CHILDES. Their work applies roughly 8,000 corrections, standardizes annotation practices across corpora, and results in the first official UD release for CHILDES speech data, which we use here to test different parsing architectures.\footnote{Concurrently to this work, a new UDPipe release has been made public at \url{https://lindat.mff.cuni.cz/services/udpipe/}, including a CHILDES\_UD-based parser that performs similarly to our Stanza parser.}


\paragraph{Structural Annotation in Acquisition Studies}
CHILDES has been an important resource for acquisition studies \citep{diessel2000development,legate2002empirical, lidz2003infants}. 
Investigations requiring structural information routinely rely on manual annotation, which is labor-intensive, thus costly and only applicable to small samples. Earlier attempts at annotating CHILDES data automatically either focus on specific phenomena (e.g., island effects, \citealp{pearl2013computational}, logical representations, \citealp{szubert2025cross}, modal verbs, \citealp{van2022figuring}, and utterance-level constructions, \citealp{bunzeck2025richness,bunzeck-etal-2025-construction}), or on very broad categories of annotation like POS tags \citep{macwhinney2008enriching,albert2013hebrew} and simplistic dependency structures \citep{sagae2010morphosyntactic}. However, most of these approaches were created \textit{ad hoc}, lack integration with standardized resources and suffer from a general mismatch between NLP tools and CHILDES data. While \citet{bunzeck2025richness} automated utterance-level construction annotation and reported a tagging accuracy of $\approx$95\%, they had to resort to a coarser construction hierarchy than previous, manually annotated studies \cite{cameron-faulkner2003construction} and focused exclusively on CDS, because results on CS were too unreliable based on the rule-based POS tags available in CHILDES at that time \citep[cf.][]{macwhinney2008enriching}. 

\section{Building a CHILDES-specific Dependency Parser and POS Tagger}

\subsection{Data}
For training CAIT, we use the annotated CHILDES dataset released by \citet{yang-etal-2025-ud}. It consists of English sentences from child--caregiver spoken interactions. The treebank contains two types of data: gold and silver. 
The gold portion consists of approximately 48k manually corrected sentences (237k words), while the silver portion is automatically annotated with \texttt{Stanza} (the combined off-the-shelf model) and covers an additional 1M sentences (6.9M words). 
The dataset is provided with predefined train, development, and test splits.
The test split contains only gold UD annotations; silver data are used exclusively for training augmentation.
The train and development splits draw from the same CHILDES corpora, while the test split comes from different corpora in order to evaluate model generalization.\footnote{For details, see \url{https://github.com/UniversalDependencies/UD_English-CHILDES}.}

\subsection{Parser Training}

We train and evaluate dependency parsers for CHILDES using four Python libraries: \texttt{Stanza} \cite{qi-etal-2020-stanza}, \texttt{SuPar}\footnote{\url{https://pypi.org/project/supar/}} \cite{zhang-etal-2020-efficient}, \texttt{DiaParser}\footnote{\href{https://github.com/Unipisa/diaparser}{\tt github.com/Unipisa/diaparser}} \cite{attardi-etal-2021-biaffine}, and \texttt{Machamp}\footnote{\href{https://github.com/machamp-nlp/machamp}{\tt github.com/machamp-nlp/machamp}} \cite{van-der-goot-etal-2021-massive}. We select these parsers for their detailed documentation, strong reported performance in dependency parsing, and prior use in parsing CHILDES by \citet{liu-prudhommeaux-2023-data}. Across these frameworks, we explore multiple architectures and training configurations to assess the impact of input representations, contextualization, and training strategies on CHILDES parsing performance.

\paragraph{Stanza} 
\texttt{Stanza} implements a BiLSTM-based deep biaffine dependency parser following \citet{dozat2017deep}. We evaluate the off-the-shelf \texttt{Stanza} (combined model \footnote{Model trained on the combination of English Treebanks: EWT, GUM, GUMReddit, PUD, Pronouns; see \url{https://stanfordnlp.github.io/stanza/combined_models.html}.}) and, separately, train parsers on UD-English-CHILDES under three input settings: (i) the default configuration using a BiLSTM encoder with the pretrained word vectors released for the CoNLL-2017 UD shared task \citep[][ named \texttt{conll17.pt}]{zeman-etal-2017-conll}; (ii) contextualized embeddings from \textit{RoBERTa-large}; and (iii) contextualized embeddings from \textit{RoBERTa-base} \cite{roberta-large}. Models were trained with a batch size of 5k for up to 50k steps, with early stopping based on development set performance using a patience of 1k steps.

\begin{table*}[t]
\centering
\small
\setlength{\tabcolsep}{5pt}
\renewcommand{\arraystretch}{1.15}
\begin{tabular}{@{}lllrrrr@{}}
\toprule
\multirow{2}{*}{Parser} & \multirow{2}{*}{Backbone Model} & \multirow{2}{*}{Training Data}
& \multicolumn{2}{c}{Development} 
& \multicolumn{2}{c}{Test} \\
\cmidrule(lr){4-5}\cmidrule(lr){6-7}
& & 
& LAS & UAS
& LAS & UAS \\
\midrule
\textbf{SuPar} & \textbf{\texttt{RoBERTa-large}} & \textbf{gold} & 93.23 & \textbf{96.23} & \textbf{92.51} & \textbf{94.91} \\
SuPar & \texttt{RoBERTa-large} & 10k silver + gold & \textbf{93.45} & 95.59 & 91.20 & 93.83 \\
SuPar & \texttt{RoBERTa-large} & 10k silver $\rightarrow$ gold & 92.31 & 94.80 & 90.41 & 93.34 \\
\midrule 
Stanza & BiLSTM & gold & 92.51 & 94.91 & 90.34 & 93.44 \\
Stanza & \texttt{RoBERTa-large} & gold & 93.28 & 95.65 & 91.39 & 94.21 \\
Stanza & \texttt{RoBERTa-base} & gold & 93.06 & 95.44 & 91.40 & 94.16 \\
Stanza & off-the-shelf & all EN-UD treebanks & 86.57 & 90.63 & 85.19 & 89.18 \\
\midrule
DiaParser & \texttt{RoBERTa-large}& gold & 91.58 & 94.48 & 89.71 & 93.03\\
DiaParser & \texttt{RoBERTa-large} & 10k silver + gold & 89.10 & 92.58 & 87.77 & 91.33\\
\midrule
Machamp & \texttt{RoBERTa-large} & gold & 85.13 & -- & 79.00 & --\\
Machamp & \texttt{RoBERTa-base} & gold & 84.46 & -- & 78.12 & --\\
\midrule 
\textit{spaCy} & \texttt{en\_core\_web\_trf} & OntoNotes 5 & 61.25 & 69.37 & 61.45 & 68.11 \\
\bottomrule
\end{tabular}
\caption{Mean LAS and UAS scores for each model on the development (3,860 sentences, 24,310 words) and test (9,591 sentences, 57,400 words) sets. Silver $\rightarrow$ gold: pretrain on silver, finetune on gold; silver~+~gold: finetune on both. Gold data refers to UD-CHILDES from \citet{yang-etal-2025-ud}. Compared with the off-the-shelf \texttt{Stanza} baseline, our best \texttt{SuPar} model's (boldface) gains on the test set are statistically significant (paired t-test over per-sentence UAS and LAS scores, 9,591 paired observations; UAS/LAS: p < 0.001).}
\label{tab:parser-results}
\end{table*}

\paragraph{SuPar}
\texttt{SuPar} implements a biaffine dependency parser supporting two encoding regimes: a \mbox{BiLSTM}\slash Transformer encoder, where token representations can be augmented with pretrained static embeddings (e.g., GloVe, \citealp{pennington-etal-2014-glove}; FastText, \citealp{bojanowski-etal-2017-enriching}) and auxiliary features such as character-level  representations and POS-tag embeddings; or a pretrained language model encoder, where token representations are obtained directly from the model without auxiliary inputs. In our experiments, we find that the latter encoding approach yields better results, and we therefore adopt \textit{RoBERTa-large} \cite{roberta-large} as the encoder backbone. We experiment with training for 10–50 epochs and observe that shorter training (around 10 epochs) leads to better generalization. 

Since \texttt{SuPar} yields the best performance in our preliminary experiments, we further explore data augmentation strategies. We consider two settings: (i) pretraining the parser on 10k silver sentences followed by fine-tuning on the gold data, and (ii) augmenting the gold training set with an additional 10k silver sentences. The silver data are obtained by re-annotating the automatically annotated silver transcripts from \citet{yang-etal-2025-ud} using our best domain-specific CHILDES parser, which we publicly release.

\paragraph{Machamp and DiaParser}
\texttt{Machamp} \cite{van-der-goot-etal-2021-massive} is a toolkit that supports multi-task learning, including dependency parsing, sequence tagging, and masked language modeling. For dependency parsing, it adopts the same deep biaffine parser of \citet{dozat2017deep} as \texttt{Stanza}. 
As for \texttt{DiaParser} \citep{attardi-etal-2021-biaffine}, it extends \texttt{SuPar} by incorporating transformer-based representations and by using attention weights from a selected transformer layer as additional features. These attention values are added to the biaffine arc-scoring function with a learned weight to provide structural hints for dependency edge prediction. However, none of the experimental settings we evaluated achieved performance comparable to \texttt{SuPar}. We report the two best-performing configurations for both parsers (Machamp and DiaParser) in \Cref{tab:parser-results}.

\paragraph{spaCy} 
Given its widespread use in dependency parsing \citep[e.g.,][]{arista2025universal,isaak2023blending}, we also include the off-the-shelf \texttt{spaCy} model \texttt{en\_core\_web\_trf}  \cite{honnibal2020spacy} as an additional baseline. \texttt{spaCy} implements a transition-based dependency parser using a variant of the non-monotonic arc-eager system \cite{honnibal-johnson-2015-improved}, trained using an imitation learning objective. Unlike the other parsers considered here, \texttt{spaCy} is not a UD-based parser. It is trained on OntoNotes 5 \cite{weischedel2011ontonotes} with ClearNLP constituent-to-dependency conversion \cite{arnold2017tidy} and does not directly produce UD-compliant dependencies; its outputs therefore require conversion before they can be compared against the UD-English-CHILDES gold standard.

\subsection{Parser Evaluation}
\label{sec:eval}
We evaluate all models using two standard dependency parsing metrics: Unlabelled Attachment Score (UAS) and Labelled Attachment Score (LAS). UAS is computed as the percentage of tokens for which the predicted head is correct, while LAS is computed as the percentage of tokens for which both the predicted head and the dependency relation label are correct. Because different parsing libraries employ distinct evaluation pipelines, direct comparison may introduce inconsistencies. 

To ensure comparability, we use a unified framework based on the \texttt{Stanza} evaluation function, which computes LAS and UAS by comparing \mbox{CoNLL-U} predictions against gold-standard annotations from the UD-English-CHILDES test set. \texttt{SuPar} and \texttt{DiaParser} natively support CoNLL-U output. 
In contrast, \texttt{spaCy}'s output is not UD-compliant and required additional pre- and post-processing to convert its outputs into valid CoNLL-U files, including remapping dependency labels to UD and correcting rare structural issues (e.g., mid-sentence index resets causing cycles or incorrect roots)\footnote{Full details on the label remapping are available in \texttt{evaluation/mapping\_parser\_spacy.py} in the repository.}.
For \texttt{Machamp}, LAS and UAS are reported using its built-in evaluation pipeline, as it does not reliably generate well-formed CoNLL-U trees compatible with the \texttt{Stanza} script.

\begin{figure*}[htb!]
\centering
\includegraphics[width=1\linewidth]{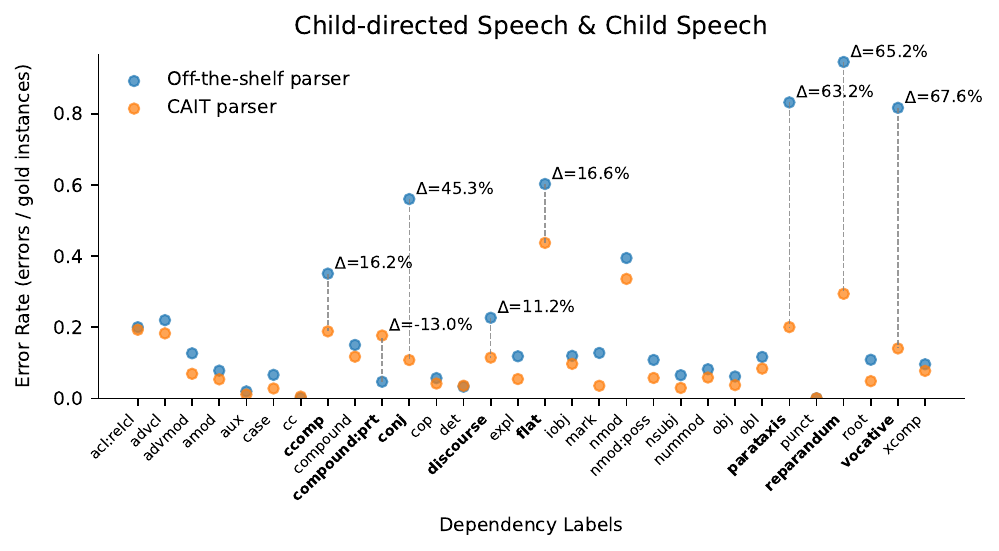}
\caption{Per-label error rates (errors normalized by gold label count) for the CAIT parser and the \texttt{Stanza} off-the-shelf baseline on the test set. Dashed lines and bold labels mark differences greater than 10 percentage points. Only dependency labels with at least 100 gold instances are included. }
\label{fig:error_comparison}
\end{figure*}

\paragraph{Results}
Table~\ref{tab:parser-results} summarizes dependency parsing performance on the development and test sets for  models trained in this study and the baselines. The best scores are obtained by a \texttt{SuPar} biaffine parser with \textit{RoBERTa-large} embeddings fine-tuned for 10 epochs, which we release as a state-of-the-art CHILDES-specific parser within CAIT and adopt as the primary reference model in subsequent analyses. \texttt{Stanza} models with contextualized embeddings are closely competitive, whereas silver data augmentation with \texttt{SuPar} does not yield consistent improvements. \texttt{DiaParser} and \texttt{MaChAmp} lag behind across all configurations and are not pursued further. All domain-specific models substantially outperform \texttt{spaCy} and the \texttt{Stanza} off-the-shelf baseline, underlining the importance of in-domain training data for parsing CHILDES.


\subsection{POS Tagger Training and Evaluation}
While our best-performing dependency parser is built on \texttt{SuPar}, this library has a notable limitation: POS tags are not used as input features during training and are therefore not predicted at inference time. As a result, the parser outputs only dependency heads and relation labels for each token, without POS annotations. To address this limitation, CAIT also includes a dedicated POS tagger trained with \texttt{Stanza}. The tagger is trained on gold data from \citet{yang-etal-2025-ud}, and can be directly loaded and applied using the standard \texttt{Stanza} pipeline, facilitating seamless integration with the CAIT parser and downstream annotation components. We evaluate the POS tagger using standard UD tagging accuracy for universal POS tags (UPOS) and language-specific POS tags (XPOS). It achieves 98.17 (UPOS) and 96.94 (XPOS) on the development set and 96.77 (UPOS) and 95.84 (XPOS) on the test set.

\section{In-Depth Analysis of \cait}
In this section, we conduct a detailed analysis of the CAIT dependency parser, in comparison with the off-the-shelf English \texttt{Stanza}. The latter serves as our baseline, as it is one of the most widely adopted tool for generating English dependency parses in current computational modeling studies \citep[e.g.,][]{kallini-etal-2024-mission,yang2026unified}. We first examine dependency relations for which the CHILDES-specific parser consistently outperforms the domain-generic baseline, highlighting cases where it mitigates systematic errors likely driven by biases in models trained on standard written corpora. We then analyze remaining error patterns, identifying phenomena for which performance gains are limited or where error rates remain relatively high. Finally, we show that CAIT can also help detect annotation inconsistencies in the gold data.

Figure~\ref{fig:error_comparison} presents a per-label comparison of parsing accuracy between the CAIT parser and the \texttt{Stanza} off-the-shelf baseline. For each dependency 
label, errors are normalized by the number of gold instances, allowing for a fair comparison across labels of varying frequency.

\subsection{Parsing Improvements}

\paragraph{\texttt{conj}}
CHILDES frequently contains repetition and bare enumeration patterns (e.g., \textit{six six seven}, \textit{purple blue green}) that are rare in standard written corpora. In the gold annotations, these are represented as flat structures with the first element as head and subsequent items attached via \texttt{conj}. The off-the-shelf parser usually wrongly analyzes these strings by selecting the final element as the \texttt{root} and attaching antecedent items as modifiers (e.g., \texttt{amod}, \texttt{nummod}). The CAIT parser instead recovers the intended flat representation (see Appendix, Figure~\ref{fig:tree_oops_dear}).


\paragraph{\texttt{vocative}}
In CHILDES, names and words of address like \textit{Thomas} or \textit{sweetheart} often appear without punctuation, unlike in typical written text. In the gold data, these are marked as \texttt{vocative} to show they are peripheral attention-getters, not part of the clause's argument structure. The off-the-shelf parser often treats them as \texttt{root} or \texttt{obj}, while the CAIT parser correctly identifies them as separate from the clause’s main dependencies (see Appendix, Figure~\ref{fig:tree_vocative}).

\begin{figure}
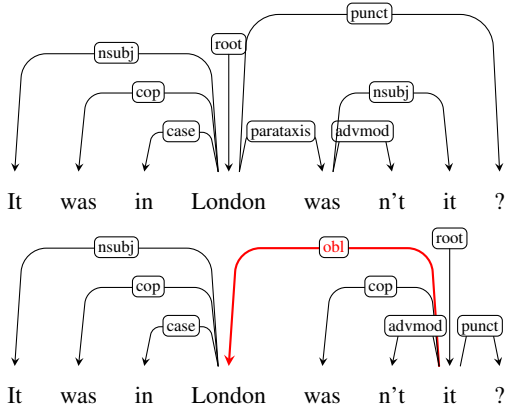

\footnotesize
    \centering
    \begin{dependency}[theme = default, edge vertical padding=2mm]
       \begin{deptext}[column sep=0.36cm]
          It \& was \& in \& London \& was \& n't \& it \& ? \\
       \end{deptext}
       \deproot{4}{root}          
       \depedge{4}{1}{nsubj}
       \depedge{4}{2}{cop}
       \depedge{4}{3}{case}
       \depedge{4}{5}{parataxis}
       \depedge{5}{6}{advmod}
       \depedge{5}{7}{nsubj}
       \depedge{4}{8}{punct}
    \end{dependency}
    \vspace{0.15cm} 
    \begin{dependency}[theme = default, edge vertical padding=2mm]
       \begin{deptext}[column sep=0.36cm]
          It \& was \& in \& London \& was \& n't \& it \& ? \\
       \end{deptext}
       \deproot{7}{root}          
       \depedge{4}{1}{nsubj}
       \depedge{4}{2}{cop}
       \depedge{4}{3}{case}
       \depedge[edge style={red, thick}]{7}{4}{\textcolor{red}{obl}}
       \depedge{7}{5}{{cop}}
       \depedge{7}{6}{advmod}
       \depedge{7}{8}{punct}
    \end{dependency}
    \caption{Gold (top) and predicted by the off-the-shelf parser (bottom) labels and relations for \texttt{parataxis}.}
    \label{fig:tree_london_prediction}
\end{figure}


\paragraph{\texttt{parataxis}}
Parataxis covers short clauses or phrases added loosely to a sentence, without being tightly connected to its main structure. In CHILDES, speakers often add questions, repeated phrases, or side comments as they speak (e.g., \textit{``…haven’t you?''}, \textit{``Excuse you excuse you''}). The off-the-shelf parser usually mislabels these by trying to fit them into the clause structure, treating them as complements (\texttt{ccomp}), copulas (\texttt{cop}) or obliques (\texttt{obl}), as in Figure~\ref{fig:tree_london_prediction}. The CAIT parser handles them better by recognizing that these extra phrases are peripheral add-ons, independent of the clause’s core structure.
\paragraph{\texttt{reparandum}}
The CAIT parser also improves performance on the \texttt{reparandum} label, which marks disfluent fragments such as stuttered segments (\textit{``N n''}) and phrase restarts (\textit{``I'm sure I'm sure...''}). The baseline parser often misclassifies these as functional categories (\texttt{det}, \texttt{cc}) or attaches them as sentence \texttt{root}, distorting the dependency structure. By correctly identifying speech repairs, the CAIT parser avoids cascading attachment errors in the rest of the tree (see Appendix, Figure~\ref{fig:tree_where_prediction}).

\subsection{Error Patterns}
Due to the domain-specific features of child–adult interaction, such as fragmentary utterances, discourse markers, and repetitions, the CAIT parser occasionally overgeneralizes certain dependency relations, resulting in a series of residual error patterns (Appendix, Figure~\ref{fig:con_matrix_cait}). Confusions involving \texttt{compound:prt} are among the most systematic errors produced by CAIT. Confusions are bidirectional between \texttt{compound:prt} and \texttt{advmod}, likely reflecting the annotation harmonization of \citet{yang-etal-2025-ud} targeting the high density of phrasal verbs in CHILDES. While \texttt{advmod} marks transparent compositional modifiers (e.g., \textit{slide down}, \textit{going around}), \texttt{compound:prt} is intended for lexicalized verb–particle units whose meaning cannot be fully predicted from their parts (e.g., \textit{hold on} `wait', \textit{wear out} `make somebody tired’). CAIT also frequently over-extends \texttt{compound:prt} to constructions such as \textit{put the pillows up}, likely because light verbs such as \textit{go}, \textit{come}, and \textit{put} appear in both literal motion uses and idiomatic phrasal verbs. Simple adverbial modifiers are sometimes misclassified as \texttt{discourse} markers, as \textit{Now} in \textit{``Now I need the blue one''}, where CAIT treats temporal words as conversational cues. Similarly, \texttt{nsubj} dependents are occasionally predicted as roots or as vocatives, reflecting CAIT’s difficulty in distinguishing between syntactic arguments and conversational address (e.g., \textit{``Who's me''} or \textit{``Mommy fix a paper''}). 


\subsection{Annotation Errors Identified by CAIT}
CAIT can also function as a diagnostic tool for detecting annotation inconsistencies when compared with a baseline such as \texttt{Stanza} trained on standard UD treebanks. Because \texttt{Stanza} largely reflects conventional UD annotation practices, systematic divergences between the two parsers may arise either from genuine performance differences or from inconsistencies in the gold annotations. We identify two annotation inconsistencies in \citet{yang-etal-2025-ud}. 

\paragraph{\texttt{det}}
The CAIT parser tends to predict possessive pronouns as \texttt{det} rather than \texttt{nmod:poss} (e.g., \textit{his coffee}, \textit{your dolly}; see Appendix, Figure~\ref{fig:con_matrix_cait}). A closer inspection of UD-English-CHILDES reveals that a small portion of possessive pronouns ($\approx$3.5\%) are annotated as \texttt{det}, deviating from standard English UD, where \texttt{det} is reserved for DET-tagged words. The parser learned this pattern from the data, making this an annotation-driven error. 

\paragraph{nmod}
Similarly, the CAIT parser shows high error rate on \texttt{nmod}, frequently predicting \texttt{compound} for premodifying noun constructions such as \textit{grape juice}, \textit{space shuttle} or \textit{tomato soup} (see Appendix, Figure~\ref{fig:tree_compound}). Per standard English UD guidelines, \texttt{compound} is correct for these constructions, while \texttt{nmod} is reserved for prepositional phrases within nominals; but in UD-English-CHILDES, $\approx$5\% of noun-premodifying-noun instances are incorrectly  annotated as \texttt{nmod}. The parser has correctly learned the standard \texttt{compound} pattern, and its elevated error rate on \texttt{nmod} thus reflects annotation noise.


Appendices~\ref{app:eval_metr}, \ref{app:er_analysis}, and \ref{app:err_patt} present additional evaluation metrics, gold–prediction tree comparisons, and a fine-grained analysis of error directions by parser type. In the next section, we leverage the CAIT parser to improve an utterance-level construction tagger and show its effectiveness in a case study.

\section{Utterance-level construction annotation}

To test the usefulness of our parser for downstream tasks in language acquisition, we build an utterance-level construction tagger on top of it, similarly to \citet{bunzeck2025richness}. Utterances are ``the primary psycholinguistic unit of child language acquisition'' \cite{tomasello2000first}. As there exists no tagger integrated with current resources and NLP tools (see Section \ref{sec:related-work}), we identify this task as an ideal opportunity to test our UD-parser. 

\paragraph{Methodology}
We approach construction tagging as a multiclass classification problem. For the classes, we devise an annotation scheme based on \citet{cameron-faulkner2003construction}: Utterances are categorized as formulaic (\texttt{FOR}, such as greetings like \textit{``Hello!''}), fragments without a predicate (\texttt{FRA}), wh- or yes/no- questions (\texttt{QWH}, \texttt{QYN}), copula sentences (\texttt{COP}, such as in Figure \ref{fig:tree_london_prediction}), imperatives (\texttt{IMP}), subject-predicate in/transitive (\texttt{SPI}, \texttt{SPT}), or complex utterances with at least two predicates (\texttt{COM}). For more details on the annotation scheme, see Appendix \ref{app:anno-guide}. We compare the following kinds of construction taggers: i) A rule-based tagger based on the UD tags provided by CAIT (cf.~Appendix \ref{app:decision-procedure}), ii) a rule-based tagger based on UD tags provided by off-the-shelf \texttt{Stanza}, iii) a rule-based tagger based only on POS tags provided by CAIT (similar to \citealp{bunzeck2025richness}), iv) a multilayer perceptron (MLP) classifier operating directly on sentence embeddings without syntactic annotation.

\paragraph{Construction tagger performance}
Table \ref{tab:cxn-results} reports performance on two different data sets. The \texttt{dev} data comes from a singular child--caregiver conversation (2,141 utterances) from the \texttt{MPI-EVA-Manchester} corpus \cite{lieven2009twoyearold}. The \texttt{test} data contains another 1,000 utterances randomly sampled from the entire CHILDES corpora to better represent potential inconsistencies in the data. All data were manually annotated according to the aforementioned annotation guide. (Appendix \ref{app:synthetic-data} contains another comparison for synthetic data.) As the MLP classifier is trained on the \texttt{dev} set, no accuracy is reported for it.

\begin{table}
\centering
\begin{tabular}{@{}lll@{}}
\toprule
Tagger & \texttt{dev} & \texttt{test} \\ \midrule
CAIT parser & 92.05 & 92.32 \\
Stanza off-the-shelf & 91.23 & 89.79 \\
POS-only & 87.54 & 85.74 \\
MLP & -- & 70.17 \\ \bottomrule
\end{tabular}
\caption{Accuracies of different tagging approaches (\%).}
\label{tab:cxn-results}
\end{table}

Across both data sets, the tagger based on our parser performs best. This holds for both CDS and CS individually (see Appendix \ref{app:speaker-type}). While the difference on the \texttt{dev} data is marginal, the differences on the \texttt{test} data are more pronounced. As the \texttt{MPI-EVA-Manchester} corpus is one of the cleanest and most standardized in CHILDES, it is not overly surprising that the standard \texttt{Stanza} tagger works reasonably well on it, but then underperforms on the \texttt{test} set that contains less systematic data (non-coded repetitions, non-standard situation descriptions, etc.). In comparison, the POS-only tagger modeled after \citet{bunzeck2025richness} yields 5--7\% worse performance, while the MLP tagger based on sentence embeddings performs dramatically worse. 

\begin{figure}[t]
\centering
\includegraphics[width=\linewidth]{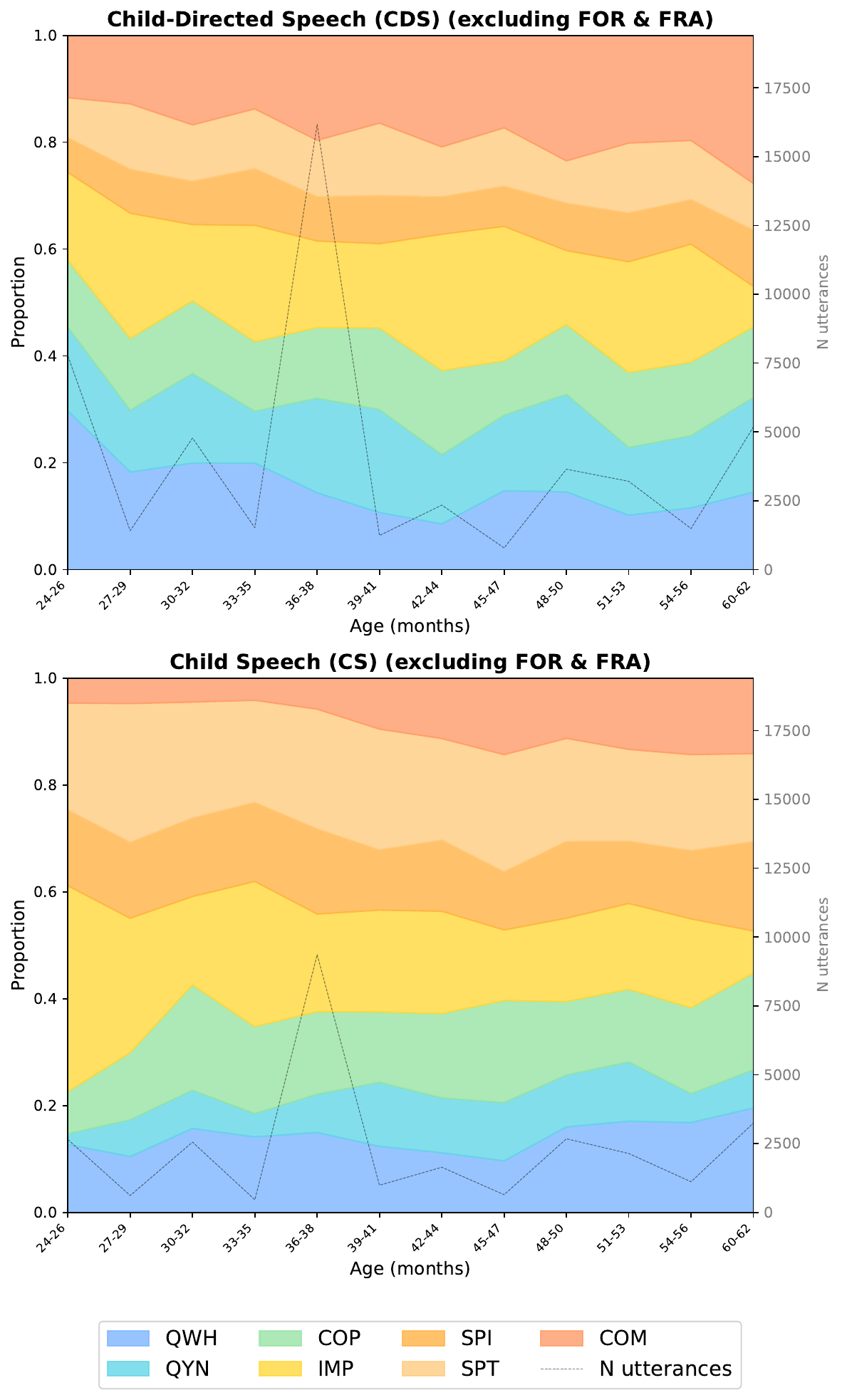}
\caption{Development of relative construction proportions in CDS and CS, measured across 3-month bins. The dashed line shows the number of utterances annotated per bin.}

\label{fig:case-study}
\end{figure}

It seems implausible that the remaining performance gap can be closed. Even with domain-specific training, certain distinctions cannot be made from text alone, due to factors such as i) archaic syntax in songs/nursery rhymes, ii)~questions and statements that differ only in prosody, iii)~run-on sentences coded as one utterance and therefore indistinguishable from complex utterances, iv)~elliptical utterances and singular verbs that cannot be clearly separated between fragments and imperatives, etc.\ (for a fine-grained analysis of specific errors, see Appendix~\ref{app:performance-analysis}).

\paragraph{Case study}

To further show the usefulness of our tagger, we use it to replicate the general methodology of \citet{bunzeck2025richness} with a portion of the \texttt{MPI-EVA-Manchester} corpus \cite{lieven2009twoyearold}. In contrast to previous studies \cite{cameron-faulkner2003construction,cameron-faulkner2011form,cameron-faulkner2013comparison,noble2018keeping,bunzeck2025richness}, however, we expand the scope considerably by including CS as well as CDS. While other studies were focused on the input only, \citet{bunzeck2025richness} focused exclusively on the development of construction types in CDS because CS was deemed too hard to parse correctly with POS tags alone. With our UD-based tagger, this is not the case. Therefore, we are able to compare the development of CDS with that of CS and complement this previous line of research. For the present case study, we exclude formulaics and fragments to focus on clausal constructions only (in line with \citealp{bunzeck2025richness}).

\citet{bunzeck2025richness} report two key findings: i) questions become less frequent in CDS across development, whereas ii) canonical SV(X) utterances become more frequent. Figure \ref{fig:case-study} shows the development of  construction types in CDS and CS between two and five years of age. Previous findings for CDS can be generally confirmed: Questions (\texttt{QWH, QYN}) in the input become less frequent when comparing the earliest age group to subsequent ones, whereas SV(X) constructions become more frequent. While standard subject--predicate utterances are fairly stable after an increase between 24 and 36 months, complex utterances see a steady increase in frequency between two and five years. Interestingly, these developments shift in CS. For the youngest age group, imperatives are most frequent. The proportion of SV(X) constructions (\texttt{SPI, SPT, COM}) increases with age, with an additional frequency shift from transitive, single-proposition sentences towards more complex utterances. Similarly, questions increase in frequency, with a clear preference for \textit{wh}-questions. These tendencies signal a shift in children's communicative behavior: Whereas the use of imperatives already allows them to communicate effectively with one-word utterances, the more complex syntax of questions emerges later. Further, the semantically prototypical agent--patient relationships expressed in transitive clauses (\texttt{SPT}) are most frequent among early propositional utterances, whereas semantically opaque constructions (e.g., complex utterances, \texttt{COM}) become more frequent with increasing age. These results should be taken with caution as they rely on a limited subset of CHILDES. Still, this case study illustrates one type of novel analysis possible with CAIT. Many more kinds of analysis are possible, such as further investigations into dialogue-oriented aspects of language development (e.g., construction-level alignment between caregivers and children, cf. \citealp{sinclair2021construction}).

\section{Conclusion \& Future Work}

We have shown that combining gold standard annotations (UD-CHILDES) with state-of-the-art parsers (\texttt{Stanza}, \texttt{SuPar}) enables reliable dependency parsing of informal, non-canonical registers like child speech and child-directed speech, which in turn can be leveraged for acquisition research, e.g., through the annotation of utterance-level constructions. With the release of our parser, POS and construction tagger we are making state-of-the-art NLP tools tailored to CHILDES data available to the acquisition community. Such UD resources are key for investigating structural properties, like dependency length, tree depth, and derived constructional measures, in light of recent arguments for the cognitive plausibility of dependency representations \cite{gibson2025syntax}, shedding further light on linguistic development and the input young learners receive. Extending this effort multilingually is a crucial next step: If even small UD-CHILDES treebanks become available in other languages, cross-lingual adaptation techniques could be employed to develop comparable parsing resources based on CAIT.

\section*{Acknowledgements}
Francesca Padovani, Jaap Jumelet and Arianna Bisazza are supported by the project `Polyglot Machines’
(VI.Vidi.221C.009) funded by the Talent Programme of the Dutch Research Council (NWO). 
Nathan Schneider is supported by NSF award IIS-2144881. 
Bastian Bunzeck is supported by the Deutsche Forschungsgemeinschaft (DFG, German Research Foundation) -- CRC-1646, project number 512393437, project A02.

\bibliography{custom,bastian,francesca}


\appendix

\section*{Limitations}

Our study is accompanied by several limitations. 

First, the CAIT toolkit assumes pre-tokenized input, as provided by UD-English-CHILDES. Researchers working with new, raw CHILDES transcripts would therefore first need a domain-specific tokenizer. While in this work we focus on the release of a POS tagger as a key component for morphosyntactic analysis, Stanza's \cite{qi-etal-2020-stanza}\footnote{\url{https://stanfordnlp.github.io/stanza/training_and_evaluation.html}} training framework provides straightforward support for training all upstream pipeline components, including tokenization (\texttt{tokenize}), multi-word token expansion (\texttt{mwt}), and lemmatization (\texttt{lemma}), directly on the gold CoNLL-U annotations of UD-English-CHILDES. Such domain-specific components could then be combined with our parser and POS tagger to form a fully end-to-end pipeline for raw, untokenized transcripts covering the full range of UD annotation layers.

Second, for our novel, CHILDES-trained UD tagger we only have the UD-CHILDES treebank available as training. While it compiles and unifies existing resources, it is still relatively small, and more data would help in improving it further. Moreover, the official UD-CHILDES treebank \citep{yang-etal-2025-ud} still contains some errors carried over from previous annotation efforts, which could inform future corrections. Some errors are likely to be not completely solvable, however, as CHILDES is not a monolithic corpus but a collection of many different corpora collected over a large time span (the English section ranges from the 1960s for the Brown corpus to the present day; the German section even contains transcripts that are over 100 years old). These corpora all adhere differently to transcription standards (e.g., pseudophonetic transcriptions like \textit{de} instead of \textit{the}), and are shaped by the affordances of the concrete theory employed or the goals of the study, reflecting \citeauthor{ochs1979transcription}'s (\citeyear{ochs1979transcription}) insight that transcription is never neutral but always a form of theoretical interpretation.

Lastly, for the construction annotation part, we concentrated on utterance-level constructions in the spirit of \citet{tomasello2000first,cameron-faulkner2003construction}, but other construction annotations should now be possible as well. The UCxn layer \citep{weissweiler2024ucxn} for UD exists, and in combination with UD-CHILDES it could open up further avenues of construction identification, especially below the utterance-level. Nevertheless, our parser already outperforms off-the-shelf approaches and could be evaluated on even more existing, developmental data. For example, on manually annotated data from \citet{diessel2000development}, the CAIT parser finds 92.1\% of relative clause constructions (537/583), whereas standard \texttt{Stanza} only finds 84.0\% (490/583). Finally, it would  also be interesting to compare our results to novel approaches to construction annotation, such as LoRA fine-tuning \citep{kaipeng2026leveraging}.

\begin{table*}[t]
\centering
\small
\setlength{\tabcolsep}{5pt}
\renewcommand{\arraystretch}{1.15}
\begin{tabular}{@{}lllrrrr@{}}
\toprule
Parser & Backbone Model & Training Data
& \multicolumn{2}{c}{Development} 
& \multicolumn{2}{c}{Test Set} \\
\cmidrule(lr){4-5}\cmidrule(lr){6-7}
& & 
& EM & UEM
& EM & UEM \\
\midrule
\textbf{SuPar} & \textbf{\texttt{RoBERTa-large}} & \textbf{gold data} & \textbf{81.17} & \textbf{89.74}&\textbf{78.22} & \textbf{87.61}\\
SuPar & \texttt{RoBERTa-large} & 10k silver data + gold data & 79.74 & 88.47 & 75.72 & 86.01 \\
SuPar & \texttt{RoBERTa-large} & 10k silver data $\rightarrow$ gold data & 80.80 & 89.61 & 77.45 &  87.40\\
\midrule 
Stanza & BiLSTM & gold data & 77.35 & 86.99 & 73.42 & 84.89 \\
Stanza & \texttt{RoBERTa-large} & gold data & 79.09 & 88.57 & 75.49 & 86.39 \\
Stanza & \texttt{RoBERTa-base} & gold data & 78.70 & 88.01 & 75.65 & 86.35 \\
Stanza & Off-the-shelf & all EN-UD treebanks & 63.78 & 77.77 & 64.07 & 76.90 \\
\midrule 
spaCy & \texttt{en\_core\_web\_trf} & OntoNotes 5 & 28.29 & 46.91 & 31.44 & 50.18 \\
\midrule
DiaParser & \texttt{RoBERTa-large}& gold data & 74.66 & 85.85 & 71.24 & 83.65 \\
DiaParser & \texttt{RoBERTa-large} & 10k silver data + gold data & 68.26 & 81.22 & 67.79 &  80.59\\
\midrule
Machamp & \texttt{RoBERTa-large} & gold data & - & - & - & - \\
Machamp & \texttt{RoBERTa-base} & gold data & - & - & - & - \\
\bottomrule
\end{tabular}
\caption{Overall EM and UEM scores for the different parsers on the development and test sets. The EM and UEM evaluation script requires grammatical tree parses and thus cannot evaluate the generated parses by Machamp.}
\label{tab:em}
\end{table*}

\section{Additional Parser Evaluation Metrics}
\label{app:eval_metr}
In Table~\ref{tab:em} we present additional sentence-level evaluation metrics for the full set of parsers discussed in the main text. Exact Match (EM) measures the proportion of sentences for which all dependency arcs and labels are correctly predicted, while Unlabeled Exact Match (UEM) considers only the correctness of heads, disregarding labels. We report overall EM and UEM scores for dev and test set, without subdividing by child speech and CDS, as we observed no substantial differences between these subsets.

Table~\ref{tab:cs-cds-results}, instead, reports the LAS and UAS breakdown by speaker role on the  test set for the CAIT parser and the \texttt{Stanza} off-the-shelf baseline. Both parsers perform better on CDS than on CS, which is consistent with the greater structural complexity in CDS and shorter, more fragmented utterances typical of child speech. The CAIT parser outperforms the baseline on both subsets, with the gap being more pronounced on CS.

\begin{table*}[t]
\centering
\small
\setlength{\tabcolsep}{5pt}
\renewcommand{\arraystretch}{1.15}
\begin{tabular}{@{}lllrrrr@{}}
\toprule
\multirow{2}{*}{Parser} & \multirow{2}{*}{Backbone Model} & \multirow{2}{*}{Training Data}
& \multicolumn{2}{c}{CS}
& \multicolumn{2}{c}{CDS} \\
\cmidrule(lr){4-5}\cmidrule(lr){6-7}
& & & LAS & UAS & LAS & UAS \\
\midrule
\textbf{SuPar} & \textbf{\texttt{RoBERTa-large}} & \textbf{gold} & \textbf{91.30} & \textbf{93.96} & \textbf{93.48} & \textbf{95.63} \\
Stanza & Off-the-shelf & all EN-UD treebanks & 83.33 & 87.81 & 86.73 & 90.31 \\
\bottomrule
\end{tabular}
\caption{CS vs.\ CDS breakdown on the test set for the best-performing trained parser (\texttt{SuPar} \texttt{RoBERTa-large}, gold) and the \texttt{Stanza} off-the-shelf baseline.}
\label{tab:cs-cds-results}
\end{table*}

Table~\ref{tab:results-by-length} reports LAS and UAS scores broken down by sentence length for both CS and CDS, comparing \texttt{Stanza} off-the-shelf and CAIT parser. Sentence length is computed excluding punctuation tokens, eventually yielding four bins: sentences of up to 3 tokens, between 4 and 6, between 7 and 10, and more than 10 tokens. The sentence counts per bin reflect well-known distributional properties of the two portions of CHILDES: CS is dominated by very short utterances, with the $\leq$3 and 4--6 bins together accounting for the large majority of sentences, while CDS exhibits a slightly higher proportion of sentences falling in the longer bins compared to CS. The CAIT parser consistently outperforms \texttt{Stanza} off-the-shelf across all bins and both speakers, with the largest gains observed on the shortest sentences in CDS, where CAIT reaches a LAS of 94.08 against \texttt{Stanza}'s 85.09. For CS, CAIT achieves its highest LAS on the 4--6 bin (92.83), while performance drops for longer sentences ($>$10), likely reflecting the scarcity of long utterances in child speech during training and the increased structural complexity they entail. Both parsers perform better on CDS than on CS across all length bins, consistent with the findings reported in Table~\ref{tab:cs-cds-results}.

\begin{table*}[t]
\centering
\small
\setlength{\tabcolsep}{5pt}
\renewcommand{\arraystretch}{1.15}
\begin{tabular}{@{}lrrrrrrrrrr@{}}
\toprule
\multirow{3}{*}{Sen. Length}
& \multicolumn{5}{c}{CS}
& \multicolumn{5}{c}{CDS} \\
\cmidrule(lr){2-6}\cmidrule(lr){7-11}
& \multirow{2}{*}{\# of Sents} & \multicolumn{2}{c}{Stanza off-the-shelf} & \multicolumn{2}{c}{CAIT parser}
& \multirow{2}{*}{\# of Sents} & \multicolumn{2}{c}{Stanza off-the-shelf} & \multicolumn{2}{c}{CAIT parser} \\
\cmidrule(lr){3-4}\cmidrule(lr){5-6}\cmidrule(lr){8-9}\cmidrule(lr){10-11}
& & LAS & UAS & LAS & UAS & & LAS & UAS & LAS & UAS \\
\midrule
$\leq$3    & 2546 & 81.03 & 86.51 & 90.11 & 93.61 & 1396 & 85.09 & 87.79 & 94.08 & 95.75\\
4--6    & 1709 & 84.71 & 88.73 & 92.83 & 95.16 & 1740 &88.63  & 92.10&94.21 & 96.43\\
7--10   & 574 & 84.37 & 88.15& 92.28 & 94.60 & 987 & 86.70&90.68 &93.10 & 95.51\\
$>$10      & 163 & 83.68 & 87.93 & 89.16& 91.57 & 476 & 85.20& 89.00& 92.02 & 94.05\\
\bottomrule
\end{tabular}
\caption{LAS and UAS by sentence length (punctuation excluded) for Child Speech (CS) and Child Directed Speech (CDS), comparing the \texttt{Stanza} off-the-shelf baseline and \texttt{SuPar} \texttt{RoBERTa-large}. \# of Sents indicates the number of sentences in each bin.}
\label{tab:results-by-length}
\end{table*}

\section{Gold and predicted trees to visualize errors}
\label{app:er_analysis}
In this section we present graphical comparisons of selected dependency trees in their gold-standard annotation and in the erroneous structures predicted by the parsers. The error category to which each example belongs is indicated in the caption and corresponds to the paragraphs discussed in the main body of the paper. The tree visualizations are intended to facilitate interpretation of the error patterns and to illustrate the structural challenges involved in accurately parsing conversational, developmental language such as CHILDES.

\begin{figure}[h]
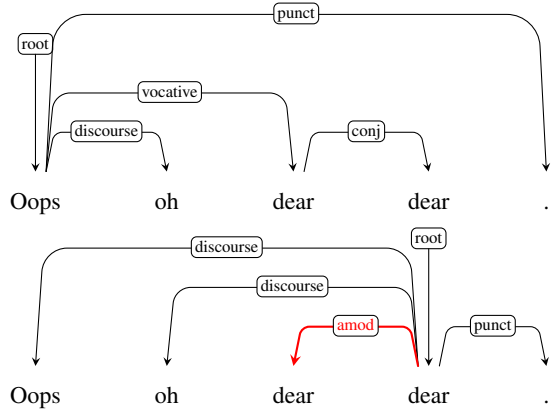

\footnotesize
    \centering

    \begin{dependency}[theme = default, edge vertical padding=2mm]
       \begin{deptext}[column sep=1.1cm]
          Oops \& oh \& dear \& dear \& . \\
       \end{deptext}
       \deproot{1}{root}
       \depedge{1}{2}{discourse}
       \depedge{1}{3}{vocative}
       \depedge{3}{4}{conj}
       \depedge{1}{5}{punct}
    \end{dependency}


    \begin{dependency}[theme = default, edge vertical padding=2mm]
       \begin{deptext}[column sep=1.1cm]
          Oops \& oh \& dear \& dear \& . \\
       \end{deptext}
       \deproot{4}{root}          
       \depedge{4}{1}{discourse}  
       \depedge{4}{2}{discourse}  
       \depedge[edge style={red, thick}]{4}{3}{\textcolor{red}{amod}} 
       \depedge{4}{5}{punct}      
    \end{dependency}

    \caption{Gold (top) and predicted by the off-the-shelf parser (bottom) dependency labels and relations (\texttt{conj}).}
    \label{fig:tree_oops_dear}
\end{figure}

\begin{figure}[h]
\footnotesize
    \centering
    \begin{dependency}[theme = default, edge vertical padding=2mm]
       \begin{deptext}[column sep=0.05cm]
          Alright \& Thomas \& let \& 's \& put \& your \& shoes \& and \& socks \& on \& . \\
       \end{deptext}
       \deproot{3}{root}
       \depedge{3}{1}{discourse}
       \depedge{3}{2}{vocative}
       \depedge{3}{4}{obj}
       \depedge{3}{5}{xcomp}
       \depedge{5}{6}{nmod:poss}
       \depedge{5}{7}{obj}
       \depedge{9}{8}{cc}
       \depedge{7}{9}{conj}
       \depedge{5}{10}{compound:prt}
       \depedge{3}{11}{punct}
    \end{dependency}
    \vspace{0.15cm}
    \begin{dependency}[theme = default, edge vertical padding=2mm]
       \begin{deptext}[column sep=0.05cm]
          Alright \& Thomas \& let \& 's \& put \& your \& shoes \& and \& socks \& on \& . \\
       \end{deptext}
       \deproot{3}{root}
       \depedge{3}{1}{discourse}
       \depedge[edge style={red, thick}]{3}{2}{\textcolor{red}{nsubj}}
       \depedge{3}{4}{obj}
       \depedge{3}{5}{xcomp}
       \depedge{5}{6}{nmod:poss}
       \depedge{5}{7}{obj}
       \depedge{9}{8}{cc}
       \depedge{7}{9}{conj}
       \depedge{5}{10}{compound:prt}
       \depedge{3}{11}{punct}
    \end{dependency}
    \caption{Gold (top) and predicted by the off-the-shelf parser (bottom) dependency labels and relations. \texttt{vocative}.}
    \label{fig:tree_vocative}
\end{figure}

\begin{figure}[h]
\footnotesize
    \centering
    \begin{dependency}[theme = default, edge vertical padding=2mm]
       \begin{deptext}[column sep=0.36cm]
          Where \& 's \& oh \& where \& the \& bus \& ? \\
       \end{deptext}
       \deproot{4}{root}
       \depedge{4}{1}{reparandum}
       \depedge{1}{2}{cop}
       \depedge{4}{3}{discourse}
       \depedge{6}{5}{det}
       \depedge{4}{6}{nsubj}
       \depedge{4}{7}{punct}
    \end{dependency}
    \vspace{0.15cm}
    \begin{dependency}[theme = default, edge vertical padding=2mm]
       \begin{deptext}[column sep=0.36cm]
          Where \& 's \& oh \& where \& the \& bus \& ? \\
       \end{deptext}
       \deproot{4}{root}
       \depedge[edge style={red, thick}]{4}{1}{\textcolor{red}{advmod}}
       \depedge{4}{2}{cop}
       \depedge{4}{3}{discourse}
       \depedge{6}{5}{det}
       \depedge{4}{6}{nsubj}
       \depedge{1}{7}{punct}
    \end{dependency}
    \caption{Gold (top) and predicted by the off-the-shelf parser (bottom) dependency labels and relations. \texttt{reparandum}.}
    \label{fig:tree_where_prediction}
\end{figure}

\begin{figure}[h]
\footnotesize
    \centering

    \begin{dependency}[theme = default]
       \begin{deptext}[column sep=1.2cm]
          What \& 's \& going \& on \& ? \\
       \end{deptext}
       \deproot{3}{root}
       \depedge{3}{1}{nsubj}
       \depedge{3}{2}{aux}
       \depedge{3}{4}{compound:prt}
       \depedge{3}{5}{punct}
    \end{dependency}

    \vspace{0.3cm} 

    \begin{dependency}[theme = default]
       \begin{deptext}[column sep=1.2cm]
          What \& 's \& going \& on \& ? \\
       \end{deptext}
       \deproot{3}{root}
       \depedge{3}{1}{nsubj}
       \depedge{3}{2}{aux}
       \depedge[edge style={red, thick}]{3}{4}{\textcolor{red}{advmod}}
       \depedge{3}{5}{punct}
    \end{dependency}
    
    \caption{Gold (top) and predicted by CAIT parser (bottom) dependency labels and relations (\texttt{compound:prt} \& \texttt{advmod}).}
    \label{fig:tree_advmod}
\end{figure}

\begin{figure}[h]
\footnotesize
    \centering

    \begin{dependency}[theme = default]
       \begin{deptext}[column sep=0.4cm]
          Are \& you \& finished \& with \& your \& juice \& ? \\
       \end{deptext}
       \deproot{3}{root}
       \depedge{3}{1}{aux:pass}
       \depedge{3}{2}{nsubj:pass}
       \depedge{3}{6}{obl}
       \depedge{6}{4}{case}
       \depedge{6}{5}{det}
       \depedge{3}{7}{punct}
    \end{dependency}

    \vspace{0.3cm} 

    \begin{dependency}[theme = default]
       \begin{deptext}[column sep=0.4cm]
          Are \& you \& finished \& with \& your \& juice \& ? \\
       \end{deptext}
       \deproot{3}{root}
       \depedge{3}{1}{cop}
       \depedge{3}{2}{nsubj}
       \depedge{3}{6}{obl}
       \depedge{6}{4}{case}
       \depedge[edge style={red, thick}]{6}{5}{\textcolor{red}{nmod:poss}}
       \depedge{3}{7}{punct}
    \end{dependency}
    
    \caption{Gold (top) and predicted by CAIT (bottom) dependency labels and relations (\texttt{det}).}
    \label{fig:tree_nmodpos}
\end{figure}

\begin{figure}[h]
\footnotesize
    \centering

    \begin{dependency}[theme = default]
       \begin{deptext}[column sep=0.5cm]
          Mom \& going \& open \& my \& toy \& box \& . \\
       \end{deptext}
       \deproot{2}{root}
       \depedge{2}{1}{nsubj}
       \depedge{2}{3}{xcomp}
       \depedge{3}{6}{obj}
       \depedge{6}{4}{nmod:poss}
       \depedge{6}{5}{nmod}
       \depedge{2}{7}{punct}
    \end{dependency}

    \vspace{0.3cm} 

    \begin{dependency}[theme = default]
       \begin{deptext}[column sep=0.5cm]
          Mom \& going \& open \& my \& toy \& box \& . \\
       \end{deptext}
       \deproot{2}{root}
       \depedge{2}{1}{vocative}
       \depedge{2}{3}{xcomp}
       \depedge{3}{6}{obj}
       \depedge{6}{4}{nmod:poss}
       \depedge[edge style={red, thick}]{6}{5}{\textcolor{red}{compound}}
       \depedge{2}{7}{punct}
    \end{dependency}
    
    \caption{Gold (top) and predicted by CAIT parser (bottom) dependency labels and relations (\texttt{nmod}).}
    \label{fig:tree_compound}
\end{figure}

\section{Error patterns for CAIT and off-the-shelf Stanza}
\label{app:err_patt}
\begin{figure*}[htb!]
\centering
\includegraphics[width=1\linewidth]{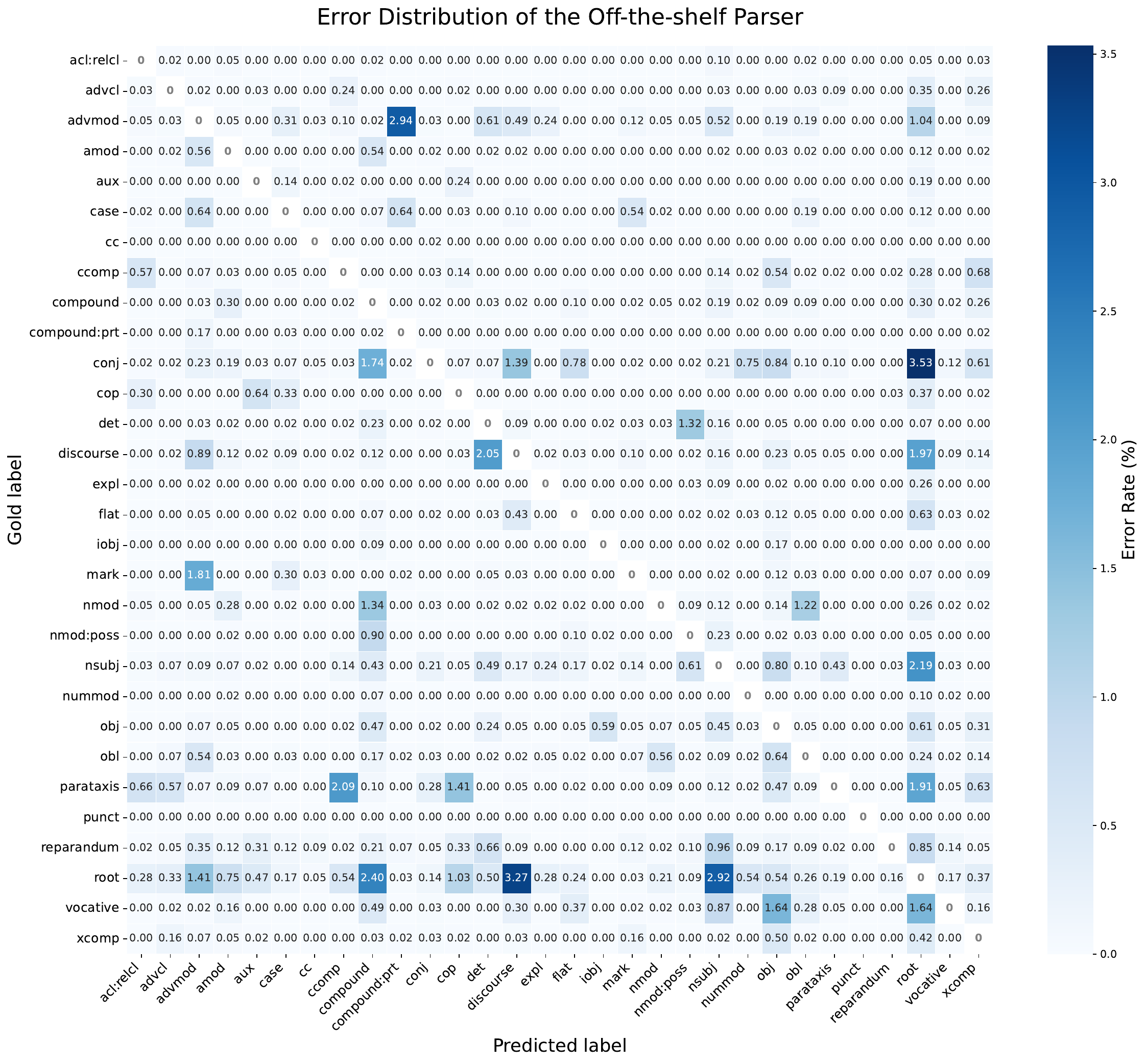}
\caption{Confusion matrix representing the error patterns of the off-the-shelf \texttt{Stanza} Parser.}
\vspace{-0.3cm} 
\label{fig:con_matrix_off}
\end{figure*}

\begin{figure*}[htb!]
\centering
\includegraphics[width=1\linewidth]{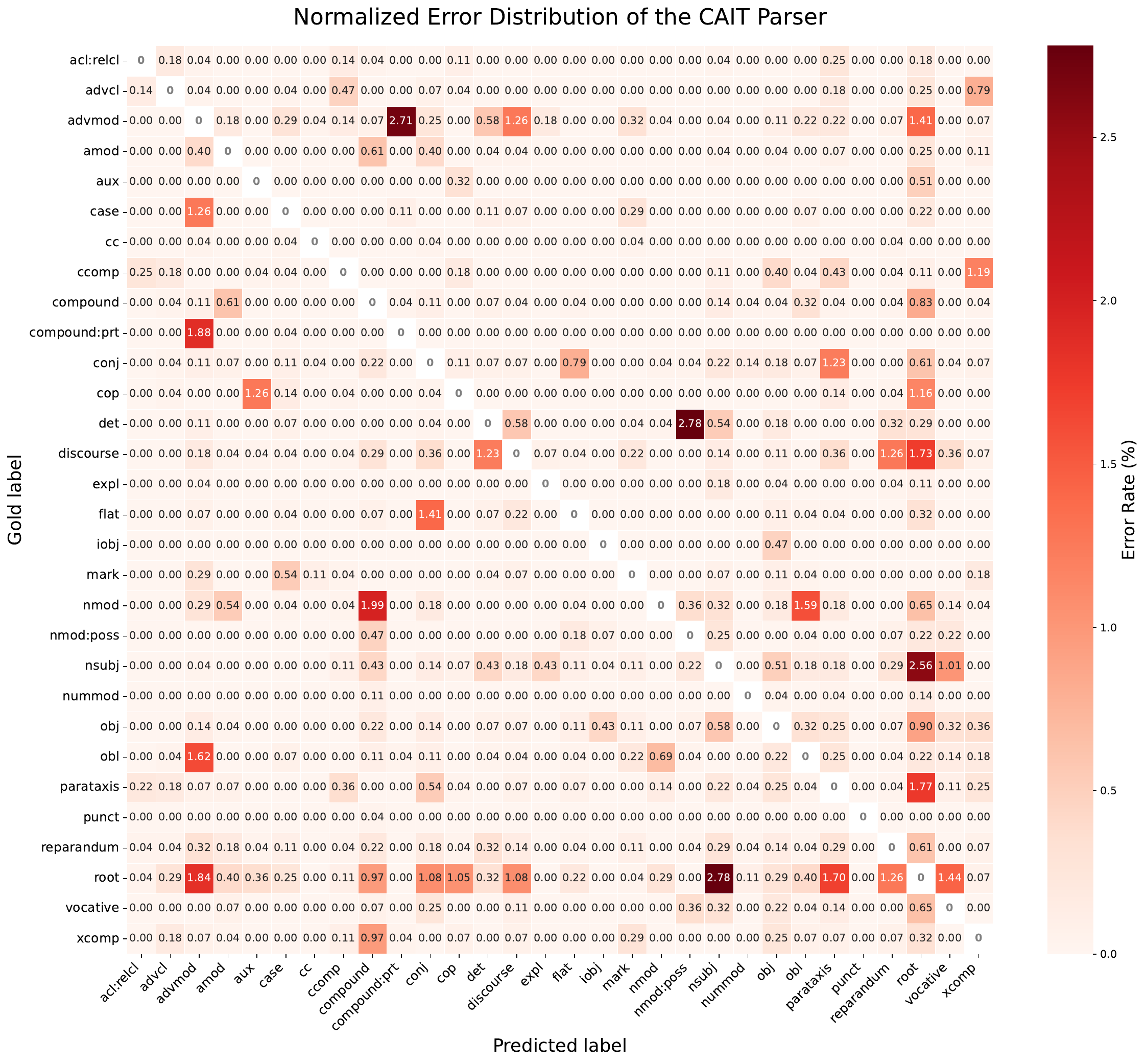}
\caption{Confusion matrix representing the error patterns of the CAIT Parser.}
\vspace{-0.3cm} 
\label{fig:con_matrix_cait}
\end{figure*}

\begin{figure*}[htb!]
\centering
\includegraphics[width=1\linewidth]{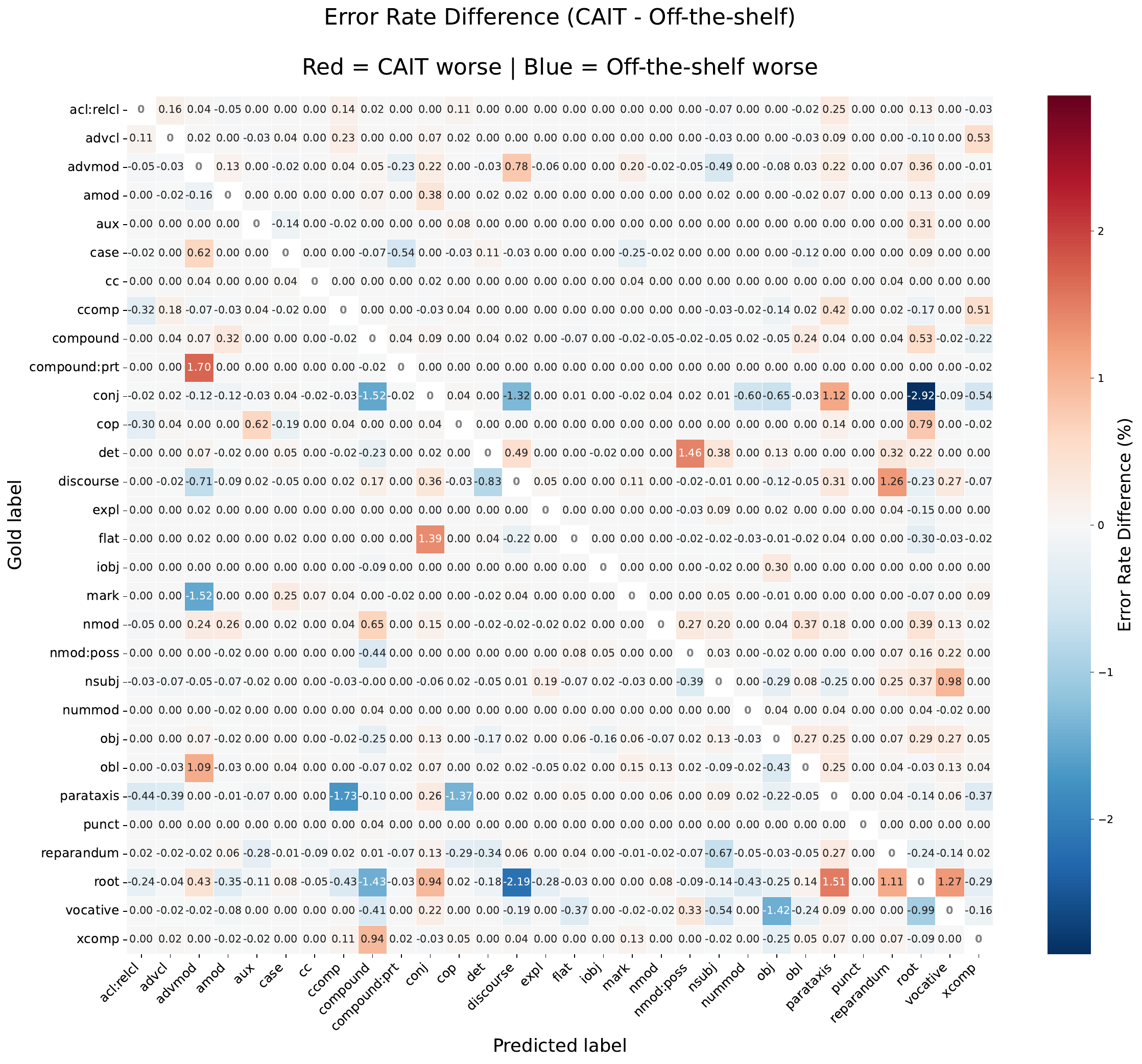}
\caption{Confusion matrix representing the delta of the error rated between the CAIT parser and the off-the-shelf \texttt{Stanza} parser. }
\vspace{-0.3cm} 
\label{fig:con_matrix}
\end{figure*}

Figures~\ref{fig:con_matrix_off}, \ref{fig:con_matrix_cait}, and \ref{fig:con_matrix} present the confusion matrices used to analyze and compare the error patterns of the two parsers: the off-the-shelf \texttt{Stanza} model and the CAIT parser.

Figures~\ref{fig:con_matrix_off} and \ref{fig:con_matrix_cait} display the normalized error distributions for the off-the-shelf \texttt{Stanza} parser and the CAIT parser, respectively. In both matrices, darker shades indicate higher misclassification rates, highlighting the most frequent confusion patterns between gold and predicted dependency labels.

Figure~\ref{fig:con_matrix} shows the difference in error rates between the two parsers (CAIT minus off-the-shelf \texttt{Stanza}). Positive values (red cells) indicate cases where CAIT produces more errors than the off-the-shelf model, while negative values (blue cells) indicate cases where the off-the-shelf model performs worse. 

\FloatBarrier

\section{Construction annotation}

\subsection{Annotation guidelines}
\label{app:anno-guide}

We devise standard construction categories in line with comparable efforts for English \cite{cameron-faulkner2003construction,cameron-faulkner2013comparison,bunzeck2025richness}, and assign one of these categories to each and every utterance. Notably, the notion of \textit{construction} employed here ranges somewhere between the traditional grammatical notion of a construction as the ``syntactic characterisation of a sentence'' \citep[1]{matthews_syntax_1996} to the more CxG conception of form-meaning pairings that occur with sufficient frequency \cite[5]{goldberg2006constructions} (cf. \citealp[4--11]{herbst2024construction}, for a comprehensive discussion of constructionhood). In sum, our \textit{constructions} are syntactically delineated, sufficiently frequent templates that serve a prototypical communicative function. We devise the following nine categories:

\begin{itemize}
    \item \texttt{FOR}: Fixed formulaic expressions (social routines): \textit{hello}, \textit{thank you}
    \item \texttt{FRA}: Fragments without finite verb: \textit{Mummy}, \textit{a baseball}
    \item \texttt{QWH}: Wh-questions (fronted interrogative): \textit{what is that?}, \textit{where are you going?}
    \item \texttt{QYN}: Yes/no questions (aux inversion, to be confirmed or denied): \textit{can you hear me?}, \textit{is it honey?}
    \item \texttt{COP}: Copula constructions: \textit{it's a kite}, \textit{the witch was green}
    \item \texttt{IMP}: Imperatives: \textit{look}, \textit{come here}, \textit{let's go}
    \item \texttt{SPI}: Subject-predicate, verb used intransitively: \textit{she laughed}, \textit{I'm going}
    \item \texttt{SPT}: Subject-predicate, verb used transitively: \textit{I love you}, \textit{she read the book}
    \item \texttt{COM}: Complex (multiple verbs/clauses, subordination or coordination pattern): \textit{I want to go and play}, \textit{the dog went home before it started raining}
\end{itemize}

Importantly, in line with \citet{cameron-faulkner2003construction}, we ignore utterance-final tag questions and remove them programmatically before the tagging process (\textit{that's good isn't it?} $\to$ \textit{that's good.}). 

\paragraph{\texttt{FOR}} 
The formulaic category is only for fixed social routine expressions. The utterance must exactly match the patterns in our tagger (ignoring punctuation and case), which consist of greetings, farewells, politeness formulaics, interjections and blessings. For a complete list please consider the tagging script in the GitHub repository. Importantly, this category requires an exact match, any additional content changes the category (\textit{oops} $\to$ \texttt{FOR}, but \textit{oops I dropped it} $\to$ \texttt{SPT}).

\paragraph{\texttt{FRA}}
Fragments are utterances that lack a finite verb or are otherwise syntactically incomplete. This includes single words (\textit{Mummy}, \textit{ball}), response particles (\textit{yeah}, \textit{mhm}), bare noun phrases (\textit{the big dog}), prepositional phrases (\textit{in there}), adjective phrases (\textit{very pretty}), incomplete copulas (\textit{she's.}, exclamatives (\textit{what a mess!}), bare participles (\textit{running}), otherwise incomplete clauses or instances of speakers trailing off (\textit{because I ...}).

\paragraph{\texttt{QWH}}
This category refers to questions introduced by a fronted interrogative pronoun (\textit{who}, \textit{what}, \textit{where}, \textit{when}, \textit{why}, \textit{how}, \textit{which}, \textit{whose}) in the main clause. The wh-word is at/near the start (before any auxiliary, ignoring additional communicatives beforehand). Although questions usually have a question mark, which is also recommended by the CHILDES guidelines, some corpora do not feature them. If it is clearly inferable from context or syntax that an utterance is meant as a question, we still do annotate it as such (and also do not strictly enforce any question mark criterion for the tagger).

\paragraph{\texttt{QYN}}
This category refers to questions with auxiliary inversion, typically expecting a yes/no answer. Crucially, we follow \citet{cameron-faulkner2003construction} in not counting sentences with declarative syntax but including a question mark (possibly indicating rising intonation) as yes/no-questions. Here structure takes precedence, although we want to stress that this remains debatable and should be critically reconsidered when adopting our construction scheme for further studies.

\paragraph{\texttt{COP}}
Copula utterances are utterances where the main predicate is a form of \textit{be} functioning as a copula (linking verb). This includes i) subject + be + predicate (\textit{it's a kite}), ii) existentials (\textit{there is a dog} and iii) passives/resultatives with \textit{be} (\textit{it's broken}, \textit{the store is closed}). Excluded are progressives like \textit{she is running} where a form of \textit{be} is an auxiliary and not a copula, and two combined copula clauses, which are considered complex utterances (\textit{it is not hot because it was rainy yesterday}).
 
\paragraph{\texttt{IMP}}
Imperatives are commands or requests, typically verb-initial. This includes bare verb commands like \textit{look}, typical subjectless phrases like \textit{come here} and negative imperatives such as \textit{don't touch it}. As mentioned, imperatives typically occur without an overt subject. Exceptions are i) hortatives or \textit{let's}-constructions (e.g., \textit{let's go}), and ii) emphatic \textit{you} imperatives (like \textit{you put that down}), which are frequent in child/child-directed language and also subsumed under imperatives due to their communicative function.

\paragraph{\texttt{SPI}, \texttt{SPT} and \texttt{COM}}
Structurally and functionally these categories are fairly similar, and \citet{bunzeck2025richness} even subsume them under one common SV(X) category. In line with \citet{cameron-faulkner2003construction}, however, we distinguish them for the sake of our analysis. All three categories require a finite verb, and their finer sub-distinction depends on i) transitivity (presence of a direct object) and ii) clause complexity (single vs. multiple predicates). 

\texttt{SPI} refers to declaratives with intransitive verbs (no direct object) like \textit{she laughed}, including progressives (\textit{she is running}) and elliptical auxiliaries (e.g., \textit{we did}). 

\texttt{SPT} refers to declaratives with transitive verbs (including direct object). This includes prototypical examples like \textit{she read the book} or \textit{I love you}. Furthermore, we include sentences that include a control verb with an infinitive complement (\textit{she needs to eat}) and utterances with objects in embedded infinitives (like \textit{I want to read that book}), as these sentences semantically still transmit a singular proposition and are structurally similar to, e.g., future constructions (cf. \textit{I will read the book} vs. \textit{I am going to read the book} vs. \textit{I want to read the book}).

\texttt{COM} then refers to utterances with multiple independent verbs or complex clausal structure. Generally, this refers to declaratives, but in the CHILDES data also combinations of other constructions are observable (like syntactic amalgams between questions and declaratives or imperatives and declaratives). We do not analyze these structures further, but subsume them under the complex category. Prototypically, this includes coordination (\textit{I want to go to the store and I will buy some milk}) or subordination patterns (\textit{I was sad because he hit me}). A general rule-of-thumb is that if two or more independent verbal predicates can be identified, then the utterance is a complex one.

\paragraph{\texttt{X}}
Finally, we used \texttt{X} as an exclusion category for a very minor share of utterances. We use it only as a last resort and annotate utterances as \texttt{X} if, and only if they i) are completely unintelligible, ii) contain only \textit{xxx}, similar transcription markers, or only metalinguistic information, or iii) adhere to non-standard annotation that have only been used for highly specific, individual corpora.

\paragraph{Decision procedure for annotation}
In general, the decision hierarchy then looks as follows:
\begin{enumerate}
    \item Is it an exact formulaic match? $\to$ \texttt{FOR}
    \item Does it lack a finite verb? $\to$ \texttt{FRA}
    \item Is it a question? Fronted wh-word in main clause? $\to$ \texttt{QWH}
    \item Is it a question? Aux inversion? $\to$ \texttt{QYN}
    \item Is the main predicate a copula (\textit{be} + ADJ/NP/passive)? $\to$ \texttt{COP}
    \item Is it a command/request (verb-initial, \textit{let's}, \textit{you} + verb)? $\to$ \texttt{IMP}
    \item  Does it have multiple verbs/clauses (coordination, subordination, etc.)? $\to$ \texttt{COM}
    \item  Does it have a direct object (includes control verb + infinitive)? $\to$ \texttt{SPT}
    \item  Otherwise $\to$ \texttt{SPI}
\end{enumerate}

\subsection{Decision procedure in tagger}
\label{app:decision-procedure}

For our UD-based tagger, we alter the previously mentioned decision procedure to a clearly delineated, progressively less strict matching algorithm:

\begin{enumerate}
\item \texttt{FOR}   $\to$ String match against formulaic patterns
\item \texttt{FRA}   $\to$ Incomplete copula (\textit{she's}) or exclamative (\textit{what a day})
\item \texttt{QYN}   $\to$ Auxiliary inversion + question mark
\item \texttt{QWH}   $\to$ Fronted wh-word in main clause
\item \texttt{COM}   $\to$ Complex clausal relations (ccomp, advcl, acl, parataxis) or conjoined verbs
\item \texttt{COP}   $\to$ Copula relation, existentials, \textit{be} + participle
\item \texttt{FRA}   $\to$ No verb root or standalone participle
\item \texttt{IMP}   $\to$ Imperative mood, \textit{you} + verb, bare verb
\item \texttt{SPI}   $\to$ Auxiliary root with subject (elliptical)
\item \texttt{SPT}   $\to$ Direct object or control verb + xcomp
\item \texttt{SPI}   $\to$ Default for remaining verbs
\item \texttt{FRA}   $\to$ Final fallback
\end{enumerate}

\subsection{Results by speaker type (CS vs. CDS)}
\label{app:speaker-type}

To further show the robustness of our tagger, we report accuracy scores divided by child-directed and child speech.

Table \ref{tab:split-accuracies-dev} shows the accuracies for the \texttt{dev} set (991 CDS utterances, 1096 CD utterances). While general accuracy tendencies hold, it can also be said that all taggers perform slightly better on CS than CDS.

\begin{table}[h]
\centering
\footnotesize
\begin{tabular}{@{}lccc@{}}
\toprule
Tagger & CDS & CS & CDS-CS $\Delta$ \\ \midrule
CAIT & 91.26\% & 93.46\% & $-$2.2\% \\
Standard \texttt{Stanza} & 88.74\% & 90.93\% & $-$2.2\% \\
POS-only & 85.44\% & 86.08\% & $-$0.6\% \\
MLP & 68.74\% & 71.73\% & $-$3.0\% \\ \bottomrule
\end{tabular}
\caption{Accuracy on \texttt{dev} data, separated by CDS vs. CS.}
\label{tab:split-accuracies-dev}
\end{table}

Table \ref{tab:split-accuracies-test} shows the accuracies for the \texttt{test} set (515 CDS utterances, 474 CS utterances).

\begin{table}[h]
\centering
\footnotesize
\begin{tabular}{@{}lccc@{}}
\toprule
Tagger & CDS & CS & CDS-CS $\Delta$ \\ \midrule
CAIT & 93.04\% & 91.15\% & $+$1.9\% \\
Standard \texttt{Stanza} & 92.03\% & 90.51\% & $+$1.5\% \\
POS-only & 88.50\% & 86.68\% & $+$1.8\% \\ \bottomrule
\end{tabular}
\caption{Tagger accuracy on \texttt{test} data, separated by CDS vs. CS.}
\label{tab:split-accuracies-test}
\end{table}

Interestingly, here the picture is slightly reversed. The taggers perform better on CDS than on CS. This suggests some remaining variation in the underlying data. As CHILDES corpora are highly diverse and annotations vary widely between different datasets, this is not overly surprising. Despite this variation, the delta between CDS and CS accuracies is approximately 2\% for all data sets, which we deem robust enough for our purposes.

\begin{table*}[htb]
\centering
\begin{tabular}{@{}lcccc@{}}
\toprule
Category & CAIT & Standard \texttt{Stanza} & POS-only & MLP with embeddings \\ \midrule
\texttt{FOR} & \textbf{98.2\%} & \textbf{98.2\%} & \textbf{98.2\%} & 67.9\% \\
\texttt{FRA} & 93.1\% & 92.4\% & \textbf{94.2\%} & 87.3\% \\
\texttt{QWH} & 97.5\% & \textbf{98.3\%} & 96.7\% & 70.2\% \\
\texttt{QYN} & 89.0\% & 89.0\% & \textbf{98.8\%} & 79.3\% \\
\texttt{COP} & \textbf{96.3\%} & 93.5\% & 93.5\% & 79.6\% \\
\texttt{IMP} & \textbf{90.7\%} & 88.8\% & 62.6\% & 69.2\% \\
\texttt{SPI} & \textbf{88.9\%} & 79.0\% & 64.2\% & 44.4\% \\
\texttt{SPT} & \textbf{84.5\%} & 83.5\% & 72.2\% & 46.4\% \\
\texttt{COM} & \textbf{90.2\%} & 73.8\% & 73.8\% & 39.3\% \\ \bottomrule
\end{tabular}
\caption{Detailed construction-tagging evaluation results for all taggers.}
\label{tab:detailed-cxn-results}
\end{table*}

\subsection{Detailed performance analysis}
\label{app:performance-analysis}

\begin{figure*}
     \begin{subfigure}[b]{0.5\textwidth}
         \includegraphics[width=\textwidth]{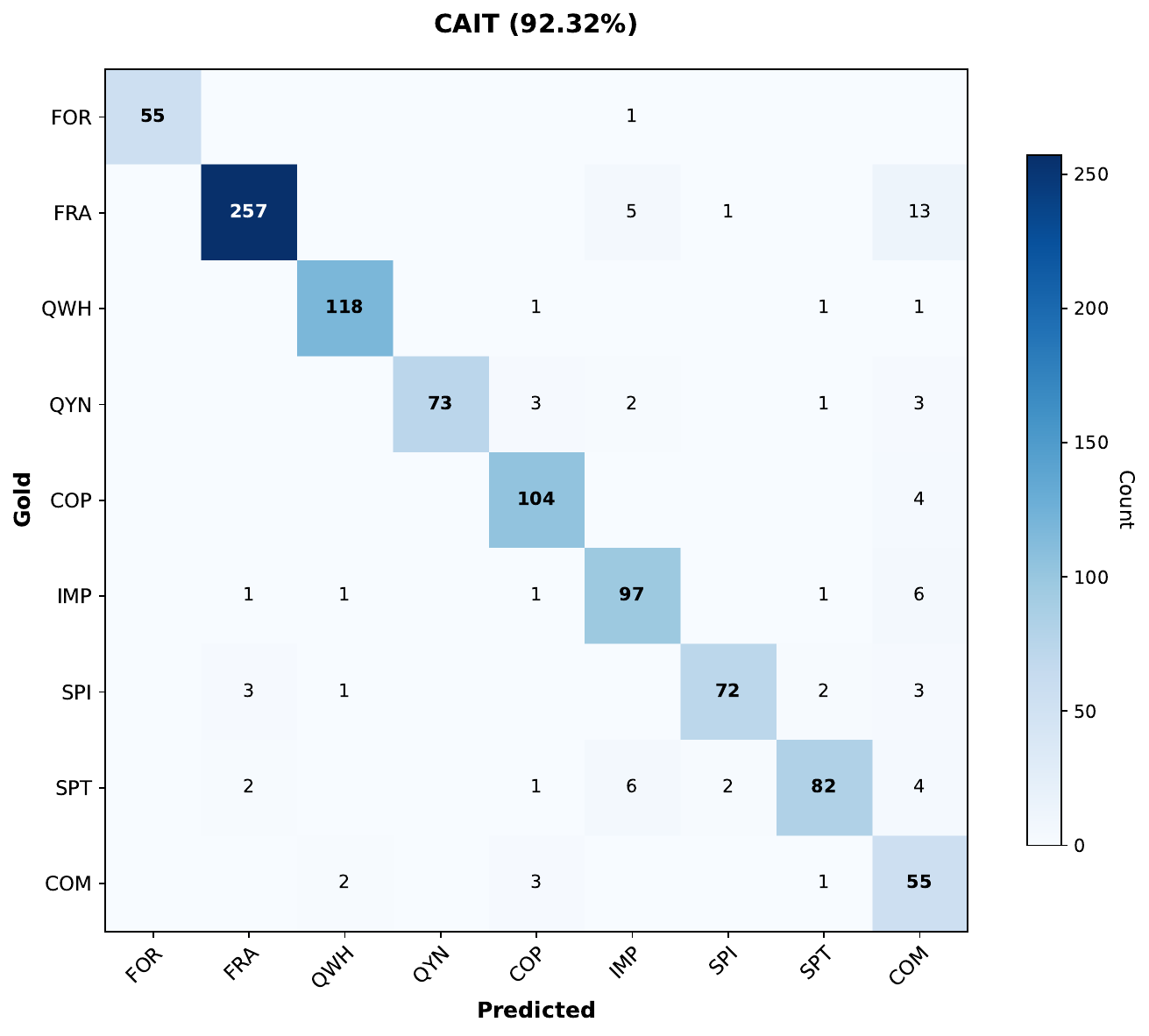}
         \caption[]{Confusion matrix for CAIT-based tagger}
         \label{fig:confusion-supar}
     \end{subfigure}
     \hfill
     \begin{subfigure}[b]{0.5\textwidth}
         \includegraphics[width=\textwidth]{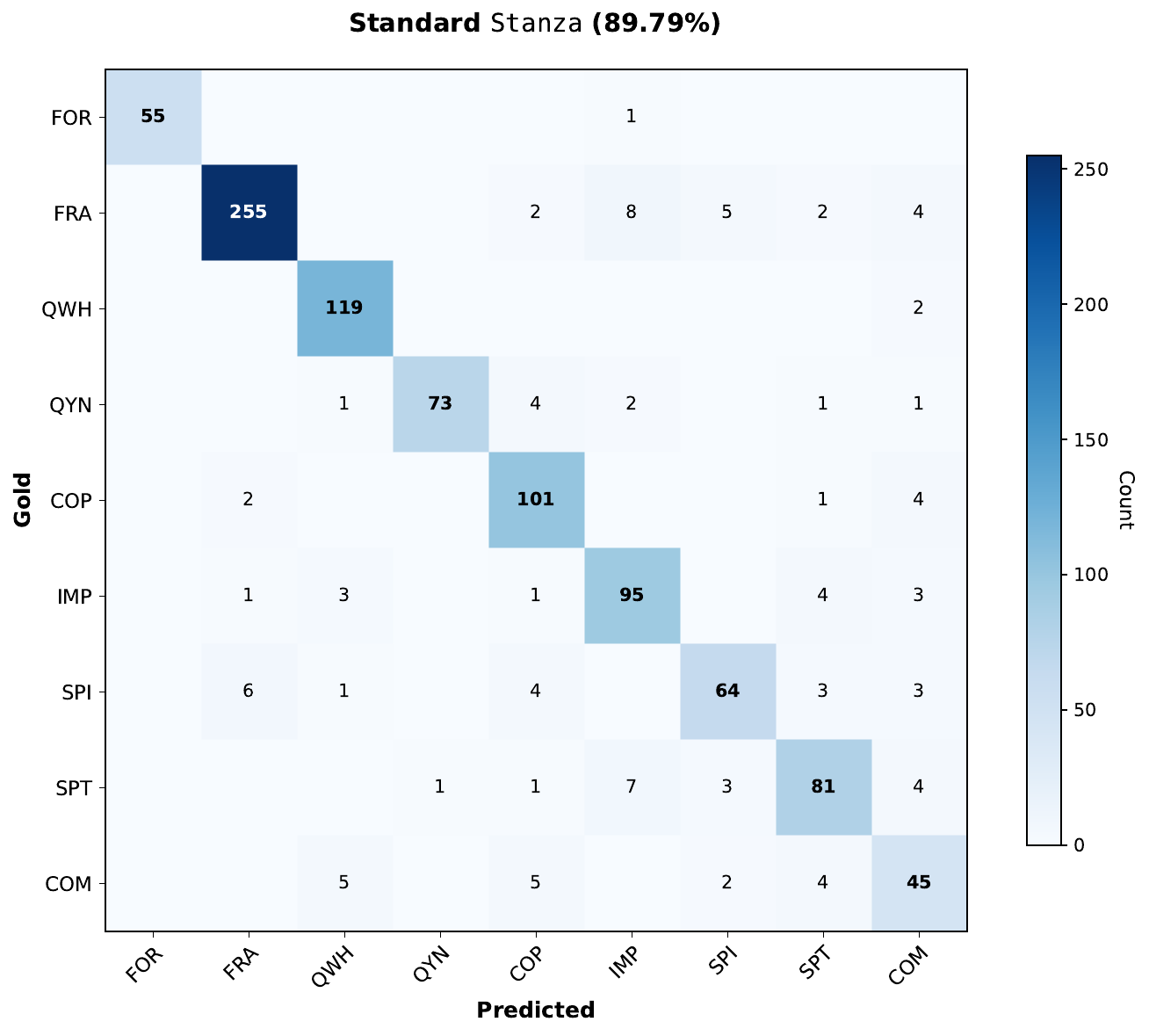}
         \caption[]{Confusion matrix, \texttt{Stanza}-based tagger} 
         \label{fig:confusion-stanza}
     \end{subfigure}

     \vskip\baselineskip
     \begin{subfigure}[b]{0.5\textwidth}
         \includegraphics[width=\textwidth]{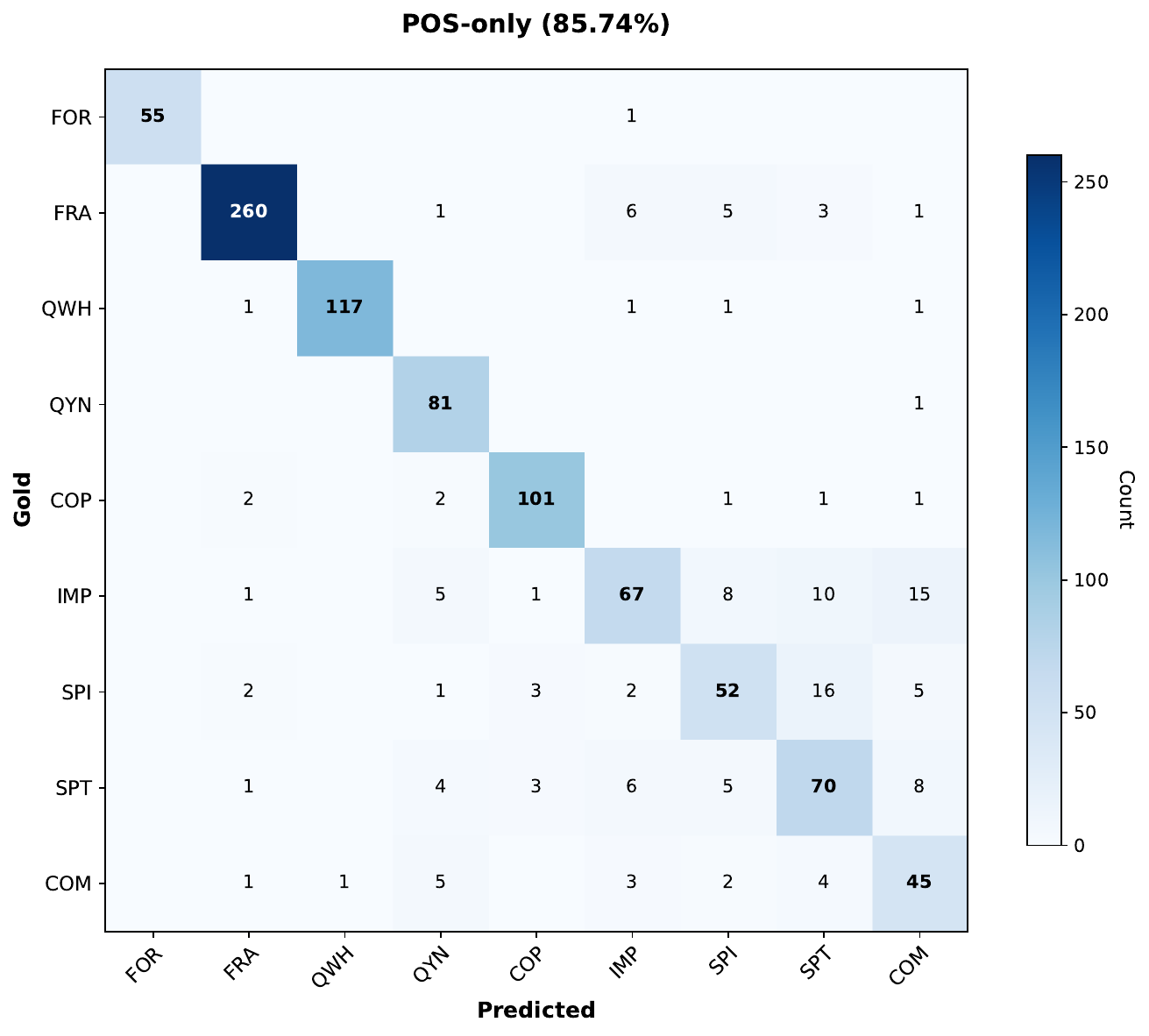}
         \caption[]{Confusion matrix, POS-based tagger} 
         \label{fig:confusion-pos}
     \end{subfigure}
     \hfill
     \begin{subfigure}[b]{0.5\textwidth}
         \centering
         \includegraphics[width=\textwidth]{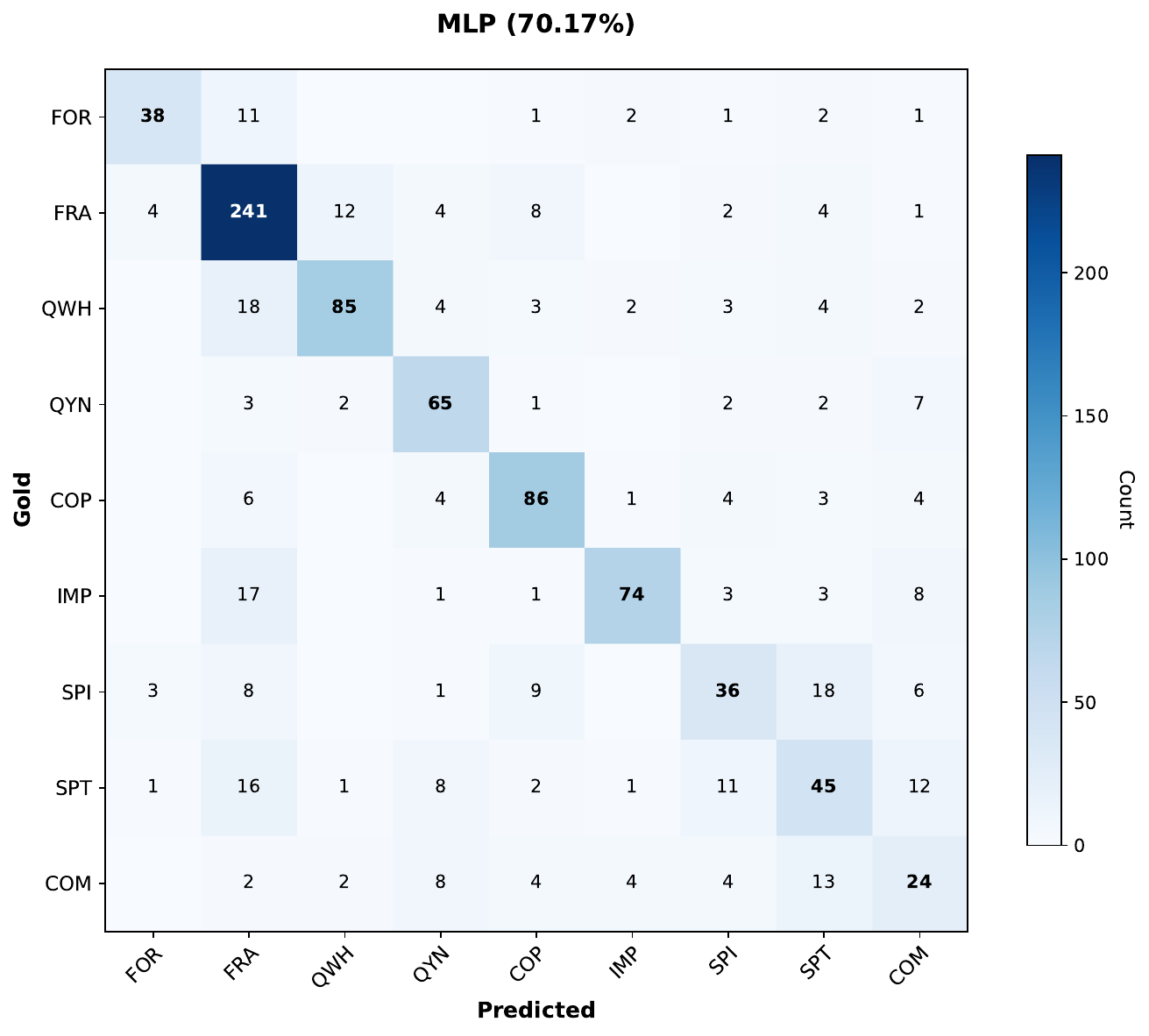}
         \caption[]{Confusion matrix, embedding-based tagger} 
         \label{fig:confusion-mlp}
     \end{subfigure}
        \caption{Confusion matrices for all construction taggers on \texttt{test} set.}
        \label{fig:confusion-matrices}
\end{figure*}

To get a more detailed overview of the advantages and shortcomings of the four individual construction taggers that we evaluated, we provide category-wise accuracies in Table \ref{tab:detailed-cxn-results} and tagger-wise confusion matrices in Figure \ref{fig:confusion-matrices}. We also report qualitative impressions that we gained from a comparison of misclassifications.

\paragraph{CAIT}
Across \texttt{test} and \texttt{dev} data, the CAIT-based tagger reaches the highest accuracy scores. The greatest improvements come from the SV(X) categories: For \texttt{SPI}, we observe an improvement of almost 10\%. The standard parser struggles with intransitive verb detection in child speech, often misclassifying \texttt{SPI} as \texttt{COP}, \texttt{FRA}, or \texttt{SPT}. CS contains many short, contextual intransitives (\textit{I'm going}, \textit{she's sleeping}) that the domain-trained parser is probably better suited to. Also for complex utterances, we notice a tremendous improvement (over 16\%). Here, the standard parser frequently fails to identify relative clauses (\texttt{acl:relcl}) and complement clauses (\texttt{ccomp}) in CHILDES data. As the dialogues often contain complex structures wrapped in simple vocabulary (\textit{that's what she said}), the mismatch between parser training data and target data between the standard \texttt{Stanza} parser and our CHILDES dialogues could be just too strong for \texttt{Stanza} to correctly recognize such structures in such simplistic and not very ``wordy'' contexts. 

In general, as Figure \ref{fig:confusion-supar} shows, there are not many confusion clusters in the CAIT-tagged data, misclassifications are spread rather evenly across the data. The only exception are fragments misclassified as complex utterances, which form a small-ish cluster. However, this might be caused by the fact that some fragments are still pretty long because they contain (unrelated) words that are strung along by the interlocutors. As dependency parsers (by design) want to assign some kind of relation, it could be that these telegraphic utterances still fulfill the criterion for complex utterances in our tagging scheme. 

\paragraph{\texttt{Stanza} off-the-shelf}
Although it slightly underperforms the CAIT-based tagger, the off-the-shelf \texttt{Stanza} tagger still reaches impressive performance across the majority of categories (except the SV(X) categories). A closer look at the confusion matrix (Figure \ref{fig:confusion-stanza}) yields that especially fragments are misclassified, also in more different ways than with the CAIT-based parser. Again, this is probably caused by the register mismatch in training data. The standard \texttt{Stanza} parser is trained on written language and long sentences, whereas spoken language features shorter and more fragmented, hesitation- and false-start-heavy data. These specific features of spoken language are then wrongly parsed and provide incorrect input to our tagging procedure.

\vspace{1mm}
\paragraph{POS-based}
Interestingly, the POS-based tagger outperforms both UD-based taggers for two categories -- fragments and yes/no-questions. On the one hand, this could be caused by the templatic nature of the POS-based tagger, which works well on the structurally rather uniform yes/no-questions. On the other hand, a closer look at the confusion matrix in Figure \ref{fig:confusion-pos} also reveals that POS-based tagger has the best recall for both fragments and yes/no-questions, but is less precise. Both categories also contain erroneous classifications that are spread out across the gold standard categories. This differs from the UD-based taggers, which are worse in terms of recall, but provide a better precision.

In contrast to \texttt{FRA} and \texttt{QYN}, the POS-based tagger works worse for \texttt{IMP} and the SV(X) categories. Imperatives are drastically under-detected. Without dependency relations, imperative detection relies on verb-initial heuristics. Many imperatives are misclassified because the POS-based tagger cannot identify the imperative mood or distinguish addressee \textit{you} from subject \textit{you}. Also, without the different types of \texttt{obj} dependency labels, object detection relies on word order heuristics. These fail for indirect objects, PP complements, and elided objects. Finally, the low accuracy on complex utterances is also caused by a lack of dependency information (subordinate clauses simply cannot be identified without \texttt{ccomp}, \texttt{advcl}, or \texttt{acl:relcl} labels).

\paragraph{Embedding-based}

The MLP classifier uses \texttt{sentence-transformers} \cite{reimers2019sentencebert} with the \texttt{all-mpnet-base-v2} embedding model to encode utterances into 768-dimensional vectors. On top of that, we train a multilayer-perceptron classifier with \texttt{scikit-learn} \cite{sklearn}. In comparison to the other classifiers, it underperforms dramatically across all categories, with the best accuracy of 87.3\% reached in the fragment category. The confusion matrix (Figure \ref{fig:confusion-mlp}) shows that misclassifications are frequent across all categories. However, this is not overly surprising, as embeddings are most useful for representing semantic features, whereas our categories are mostly syntactically defined. The 22\% between embedding and UD-based parsing demonstrates that such kinds of syntactic classification require explicit structural analysis. Again, this becomes most apparent for the SV(X) categories, where \texttt{SPI} and \texttt{SPT} are frequently confused. Similarly, complex utterances are classified into the whole spectrum of construction types (except formulaics), showing that embeddings hardly represent clause structure (for example, \textit{I think he's happy} and \textit{he's happy} should embed similarly, despite their important structural differences).

\begin{figure*}[htb!]
\centering
\includegraphics[width=0.8\linewidth]{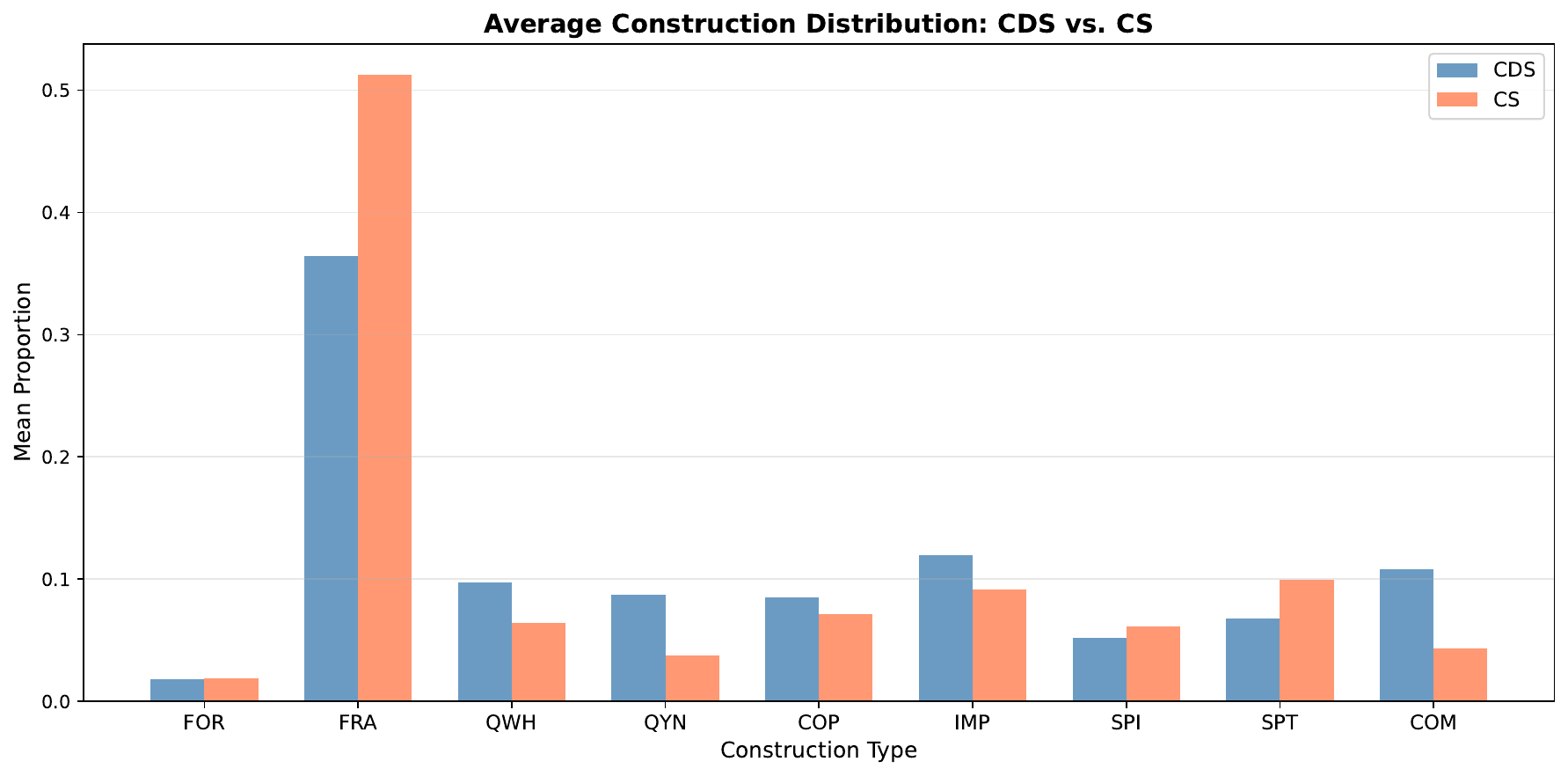}
\caption{Mean proportion of construction types across \texttt{MPI-EVA-Manchester} corpus sample.}
\vspace{-0.3cm} 
\label{fig:average-dist}
\end{figure*}

\subsection{Synthetic data comparison}
\label{app:synthetic-data}

To further assess tagger generalization, we evaluated on a synthetic dataset of 800 programmatically generated utterances (100 per category, excluding FOR). We generated these utterances by prompting  \texttt{gpt-oss-20b} \citep{openai2025gptoss120b} to produce 100 sentences that fit a short category description and contain language that is adequate for children. They were originally used for creating a very first tagging schema, but then abandoned in favor of the \texttt{dev} data set. For the sake of comparison we decided to assess tagger performance on the synthetic data after the development of all taggers was finished.

\begin{table}[htb!]
\centering
\begin{tabular}{@{}ll@{}}
\toprule
Tagger & Accuracy \\ \midrule
CAIT & 92.00\% \\
Standard \texttt{Stanza} & 91.87\% \\
POS-only & \textbf{92.37\%} \\
MLP & 55.75\% \\ \bottomrule
\end{tabular}
\caption{Tagger evaluation results for synthetic data.}
\label{tab:synthetic-data-results}
\end{table}

Table \ref{tab:synthetic-data-results} shows a somewhat different picture compared to the naturally occurring datasets. The POS-based tagger outperforms the UD-based taggers (by a minimal margin). This might be due to the syntactic uniformity of synthetic data, which deviates quite drastically from naturally occurring data (cf. \citealp{ju2025domain}). As there is little variation and deviation in its patterns, the purely pattern-driven POS tagger has no problems identifying the constructions in the correct way. Interestingly, the embedding classifiers collapse completely. It is plausible that the purely distributional patterns that they learn from CHILDES data do not generalize to artificially constructed data.

\subsection{Further results for case study}

Figure \ref{fig:average-dist} displays the mean proportion of construction types across the complete \texttt{MPI-EVA-Manchester} corpus sample, regardless of age. Here, the contrasts between CDS and child speech become most visible. Fragments and (in)transitive propositional utterances are more frequent in child speech than in CDS; whereas questions, copulas, imperatives and complex utterances occur more in CDS than in child speech. From a syntactic viewpoint, this contrast is somewhat expected, as the sample is taken from young children whose mental constructicon is still more item-based and less abstract/schematic. On the semantic/discourse-pragmatic level, the interaction of these categories should also be considered. Children are not linguistic mirrors, but engage in conversation with their caregivers. Questions in CDS beget answer in child speech, e.g., in the form of one-word fragments or simple propositions. 

\begin{figure*}[htb!]
\centering
\includegraphics[width=0.8\linewidth]{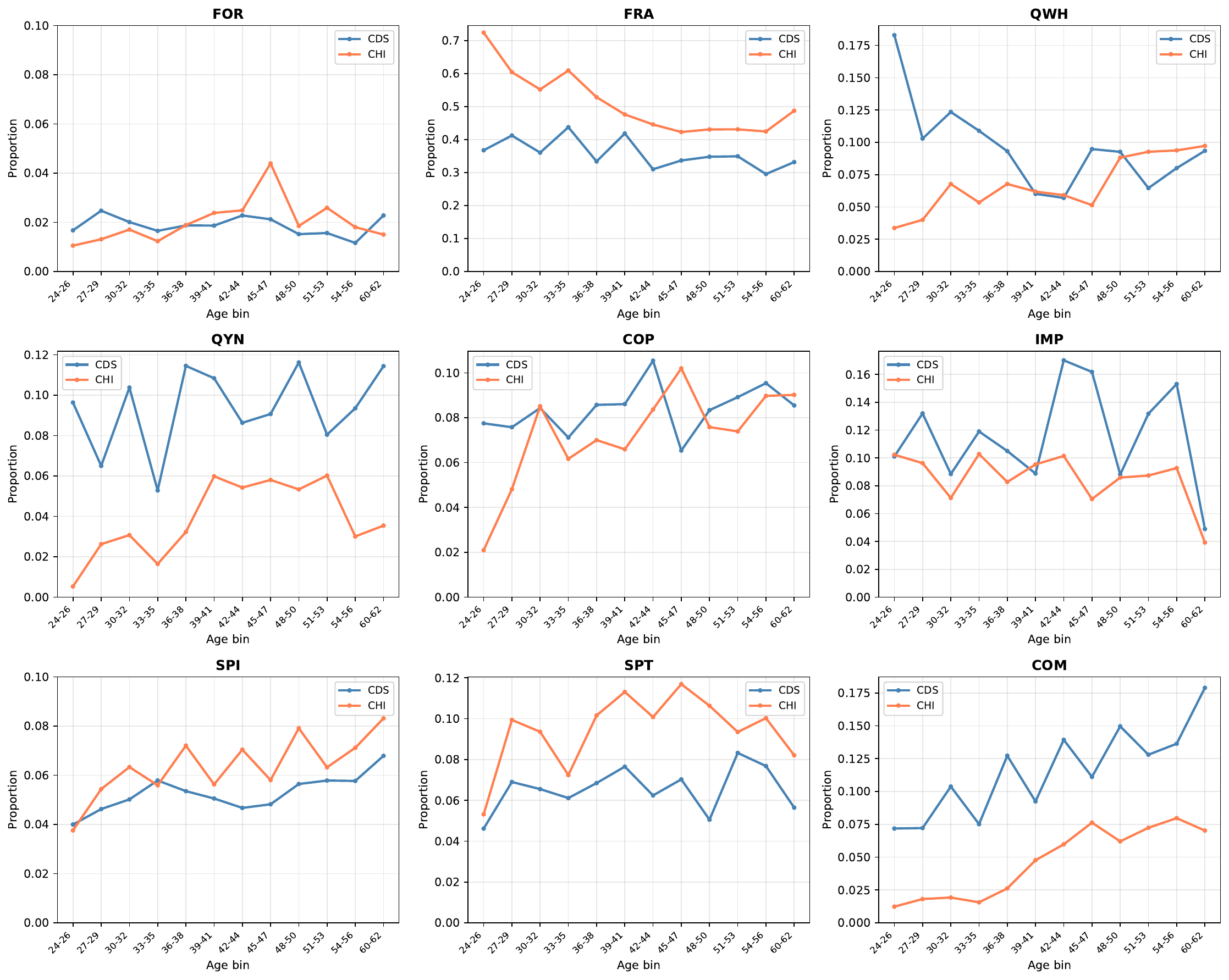}
\caption{Development of individual construction-type frequencies across dataset, separated by speaker (caregiver vs. child).}
\vspace{-0.3cm} 
\label{fig:cds-vs-chi}
\end{figure*}

Figure \ref{fig:cds-vs-chi} shows the development of construction type frequency across time, divided by CDS vs. CS. The most interesting difference here can be found in the development of wh-question frequency. Wh-questions are the only construction type where the CDS and the child speech development curves are almost mirrored. From an interaction perspective, however, this is not overly surprising. In early acquisition, caregivers try to elicit answers from children and motivate communicative behavior. This is simple with wh-questions, which can already be answered with one-word fragments (and also actively teach novel vocabulary). When this first bootstrapping period is done, the initiative in communication shifts to children who want to know things about the world they live in and actively seek out new words themselves by requesting them.

\begin{figure*}[htb!]
\centering
\includegraphics[width=0.8\linewidth]{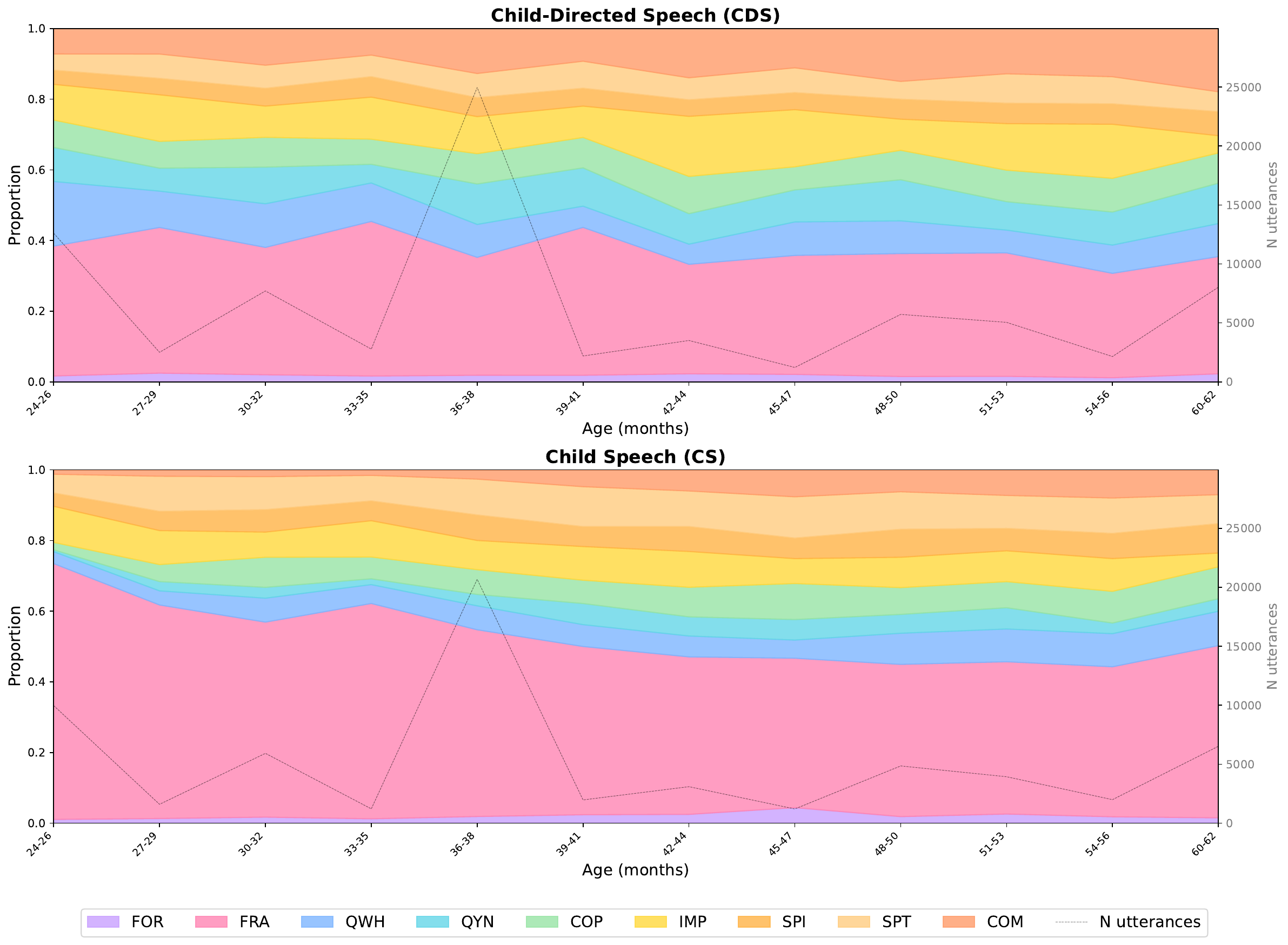}
\caption{Development of construction type frequencies in child-directed and child speech (including \texttt{FOR} and \texttt{FRA}).}
\vspace{-0.3cm} 
\label{fig:case-study-full}
\end{figure*}

Finally, Figure \ref{fig:case-study-full} shows the proportions also portrayed in Figure \ref{fig:case-study}, but including syntactic fragments and formulaics. In CDS, the frequency of fragments is not dramatically impacted by child age, and formulaics stay highly infrequent across development. In contrast, CS features almost 80\% fragments in the first age group, which then steadily decreases towards the 40--50\% range. Formulaics and social routines become frequent around 4--5 years of age.

\end{document}